\documentclass[10pt,journal,compsoc,twoside]{IEEEtran}

\usepackage{times}
\usepackage{epsfig}
\usepackage{graphicx}
\usepackage{amsmath}
\usepackage{amssymb}

\usepackage{algorithm}
\usepackage{algorithmic}
\usepackage{caption}
\usepackage{bm}

\usepackage{color}
\usepackage{hyperref}

\usepackage{amsmath}

\ifCLASSOPTIONcompsoc
\usepackage[nocompress]{cite}
\else
\usepackage{cite}
\fi

\ifCLASSINFOpdf
\else
\fi

\begin{document}
\title{Dynamic Scene Deblurring using \\
	a Locally Adaptive Linear Blur Model}

\author{Tae Hyun Kim, Seungjun Nah, and Kyoung Mu Lee
	\IEEEcompsocitemizethanks{\IEEEcompsocthanksitem The authors are with the Department of Electrical Engineering and
		Computer Science, Automation and Systems Research Institute, Seoul
		National University, 1 Ganak-ro, Gwanak-gu, Seoul 151-744, South
		Korea (E-mail: lliger9@snu.ac.kr, seungjun.nah@gmail.com, kyoungmu@snu.ac.kr).}
}

%
%

\markboth{~}
{Kim \MakeLowercase{\textit{et al.}}: Dynamic Scene Deblurring using a Locally Adaptive Linear Blur Model}
%



\IEEEtitleabstractindextext{%
\begin{abstract}
State-of-the-art video deblurring methods cannot handle blurry videos recorded in dynamic scenes, since they are built under a strong assumption that the captured scenes are static.
Contrary to the existing methods, we propose a video deblurring algorithm that can deal with general blurs inherent in dynamic scenes.
To handle general and locally varying blurs caused by various sources, such as moving objects, camera shake, depth variation, and defocus, we estimate pixel-wise non-uniform blur kernels.
We infer bidirectional optical flows to handle motion blurs, and also
estimate Gaussian blur maps to remove optical blur from defocus in our new blur model.
Therefore, we propose a single energy model that jointly estimates optical flows, defocus blur maps and latent frames.
We also provide a framework and efficient solvers to minimize the proposed energy model.
By optimizing the energy model, we achieve significant improvements in removing general blurs, 
estimating optical flows, and extending depth-of-field in blurry frames.
Moreover, in this work, to evaluate the performance of non-uniform deblurring methods objectively, we have constructed a new realistic dataset with ground truths.
In addition, extensive experimental on publicly available challenging video data demonstrate that the proposed method produces qualitatively superior performance than the state-of-the-art methods which often fail in either deblurring or optical flow estimation.

\end{abstract}

\begin{IEEEkeywords}
video deblurring, non-uniform blur, motion blur, defocus blur, optical flow
\end{IEEEkeywords}}

\maketitle

\IEEEdisplaynontitleabstractindextext

%
\IEEEpeerreviewmaketitle

\section{Introduction}

\IEEEPARstart{M}{otion} blurs are the most common artifacts in videos recorded from hand-held cameras.
In low-light conditions, these blurs are caused by camera shake and object motions during exposure time.
In addition, fast moving objects in the scene cause blurring artifacts in a video even when the light conditions are acceptable.
For decades, this problem has motivated considerable works on deblurring and
different approaches have been sought depending on whether the captured scenes are static or dynamic.

Early works on a single image deblurring problem are based on assumptions that the captured scene is static and has constant depth~\cite{Cho:2009,Fergus:2006,Gupta:2010,Hirsch:2011,Shan:2008,Whyte:2012} and they estimated uniform or non-uniform blur kernel by camera shake.
These approaches were naturally extended to video deblurring methods.
Cai et al.~\cite{cai2009blind} proposed a deconvolution method with multiple frames
using sparsity of both blur kernels and clear images to reduce errors from inaccurate registration and render high-quality latent image.
However, this approach removes only uniform blur caused by two-dimensional translational camera motion,
and the proposed approach cannot handle non-uniform blur from rotational camera motion around z-axis,
which is the main cause of motion blurs~\cite{Whyte:2012}.
To solve this problem, Li et al.~\cite{li2010generating} adopted a method 
parameterizing spatially varying motions with 3x3 homographies based on the previous work of Tai et al.~\cite{tai2011richardson}, and could handle non-uniform blurs by rotational camera shake.
In the work of Cho et al.~\cite{cho2012registration}, camera motion in three-dimensional space was estimated
without any assistance of specialized hardware, and spatially varying blurs caused by projective camera motion were obtained.
Moreover, in the works of Paramanand et al.~\cite{paramanand2013non} and Lee and Lee~\cite{lee2013dense}, spatially varying blurs by depth variation in a static scene were estimated and removed.

\begin{figure*}[t]
\begin{center}
\includegraphics[width=\linewidth]{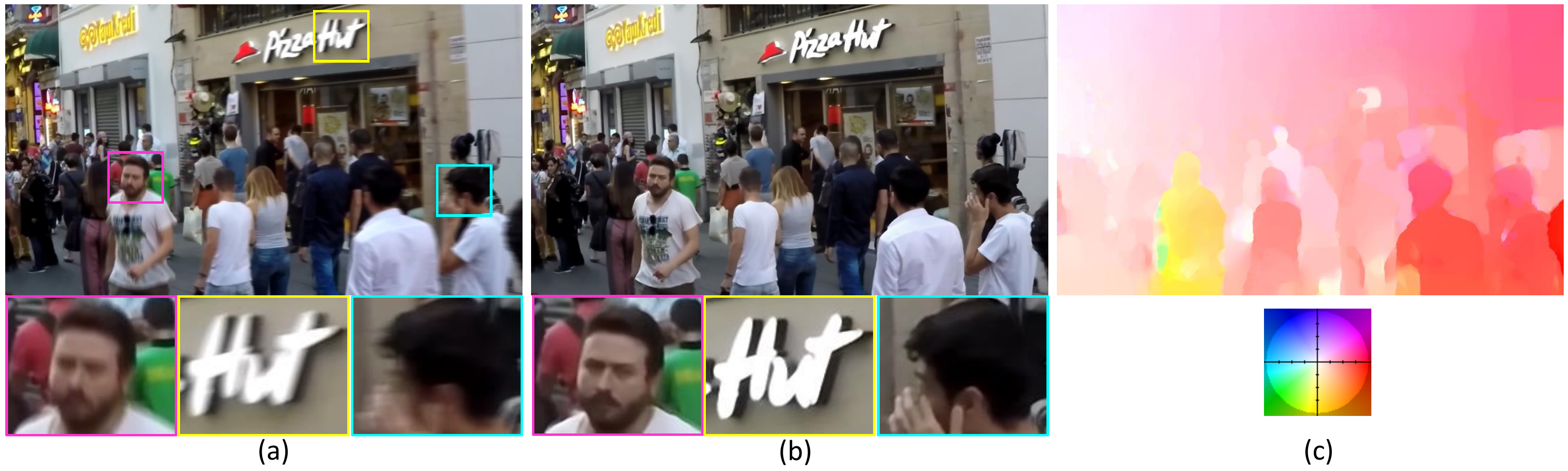}
\end{center}
\caption{(a) A Blurry frame in a dynamic scene. (b) Our deblurring result. (c) Our color coded optical flow estimation result.}
\label{fig_intro}
\end{figure*}

However, these previous methods, which assume static scene,
suffer from spatially varying blurs from not only camera shake but also moving objects in a dynamic scene.
Because it is difficult to parameterize the pixel-wise varying blur kernel in the dynamic scene with simple homography, kernel estimation becomes more challenging task.
Therefore, several researchers have studied on removing blurs in dynamic scenes,
which are grouped into two approaches: segmentation-based deblurring approach, and exemplar-based deblurring approach.

Segmentation-based approaches usually estimate multiple motions, kernels, and associated segments.
In the work of Cho et al.~\cite{cho2007removing}, a method 
that segments homogeneous motions and estimates segment-wise different 1D Gaussian blur kernels, was proposed.
However, it cannot handle complex motions by rotational camera shakes due to the limitation of Gaussian kernels.
In the work of Bar et al.~\cite{bar2007variational}, a layered model was proposed
that segments images into foreground and background layers,
and estimates a linear blur kernel within the foreground layer.
By using the layered model, explicit occlusion handling is possible,
but the kernel is restricted to linear.
To overcome these limitations, 
Wulff and Black~\cite{Wulff:ECCV:2014} improved the previous layered model of Bar et al.
by estimating the different motions of both foreground and background layers.
However, these motions are restricted to affine models
and it is difficult to extend to multi-layered scenes because such task requires depth ordering of the layers.
To sum up, segmentation-based deblurring approaches have the advantage of removing blurs caused by moving objects in dynamic scenes. 
However, segmentation itself is very difficult problem and remains still an challenging issue as reported in~\cite{thkim_cvpr2014}.
Moreover, they fail to segment complex motions like motions of people,
because simple parametric motion models used in~\cite{bar2007variational,Wulff:ECCV:2014} 
cannot fit the complex motions accurately.

Exemplar-based approaches were proposed in the works of Matsushita et al.~\cite{matsushita2006full} 
and Cho et al.~\cite{cho_siggraph2012}.
These methods usually do not rely on accurate segmentation and deconvolution.
Instead, the latent frames are rendered by interpolating lucky sharp frames that frequently exist in videos, thus avoiding  severe ringing artifacts.
However, the work of Matsushita et al.~\cite{matsushita2006full} cannot remove blurs caused by moving objects.
In addition the work of Cho et al.~\cite{cho_siggraph2012} allows only slow-moving objects in dynamic scenes
because it searches sharp patches corresponding to blurry patch after registration with homography.
Therefore, it cannot handle fast moving objects which have distinct motions from those of backgrounds.
Moreover, since it does not use deconvolution
with spatial priors but simple interpolation, it degrades mid-frequency textures such as grasses and trees, and renders smooth results.

On the other hand, defocus from limited depth-of-field (DOF) of conventional digital cameras also results in blurry effects in videos. Although shallow DOF is often used to render aesthetic images and highlight the focused objects, frequent misfocus of moving objects in video yields image degradation when the motion is large and fast. Moreover, depth variation in the scene generates spatially varying defocus blurs, making the estimation of defocus blur map is also a difficult problem. Thus many researches have studied to estimate defocus blur kernel. Most of them have approximated the kernel as simple Gaussian or disc model,
making the kernel estimation problem becomes a parameter (e.g. standard deviation of Gaussian blur, disc radius) estimation problem~\cite{bae2007defocus,kee2011modeling,zhu2013estimating,zhuo2011defocus}.

To magnify focus differences, Bae and Durand~\cite{bae2007defocus} estimated defocus blur map at the edges first, and then propagated the results to other regions. However, the estimated blur map is inaccurate where the blurs are strong, since it is image-based approach and depends on the detected edges that can be localized.
Similarly, Zhuo and Sim~\cite{zhuo2011defocus} propagated the amount of blur at the edges to elsewhere,
that obtained by measuring the ratio between the gradients of the defocused input and re-blurred input with a Gaussian kernel. 
To reduce reliance on strong edges in the defocused image, Zhu et al.~\cite{zhu2013estimating} utilized 
statistics of blur spectrum within the defocused image, 
since statistical models could be applicable where there are no strong edges.
Specifically, local image statistics is used to measure the probability of defocus scale and determine the locally varying scale of defocus blur in a single image. 
However, local image statistics-based methods do not work when there are motion blurs as well as defocus blurs within a single image; Motion blurs change local statistics and yield much complex blurs combined with defocus blurs.

In the recent work of Kim and Lee~\cite{thkim_cvpr2015}, 
a new and generalized video deblurring (GVD) method
that estimates latent frames without using global motion parametrization and segmentation was proposed 
to remove motion blurs in dynamic scenes.
In GVD, bidirectional optical flows are estimated and used to infer pixel-wise varying kernels.
Therefore, the proposed method naturally handle coexisting blurs by camera shake, and moving objects with complex motions. Because estimating flow fields and restoring sharp frames are a joint problem, both variables are simultaneously estimated in GVD. 
To do so, a new single energy model to solve the joint problem was proposed and efficient solvers to optimize the model is provided.

However, since GVD method is based on piece-wise linear kernel approximation, it cannot handle non-linear blurs combined with motion and defocus blurs which are common in videos captured from hand-held cameras.
Therefore, in this work, we propose an extended and more generalized method of GVD that can handle not only motion blur but also defocus blur
which further improves the deblurring quality significantly. 
Under an assumption that, the complex non-linear blur kernel can be decomposed into motion and defocus blur kernels, we estimate bidirectional optical flows to approximate motion blur kernel, scales of Gaussian blurs to approximate defocus blur kernel, and the latent frames jointly. 
The result of our system is shown in Fig.\ref{fig_intro},
in which the motion blurs of differently moving people and Gaussian blurs in the background are successfully removed and accurate optical flows are jointly estimated.

\begin{figure*}[t]
	\begin{center}
		\includegraphics[width=\linewidth]{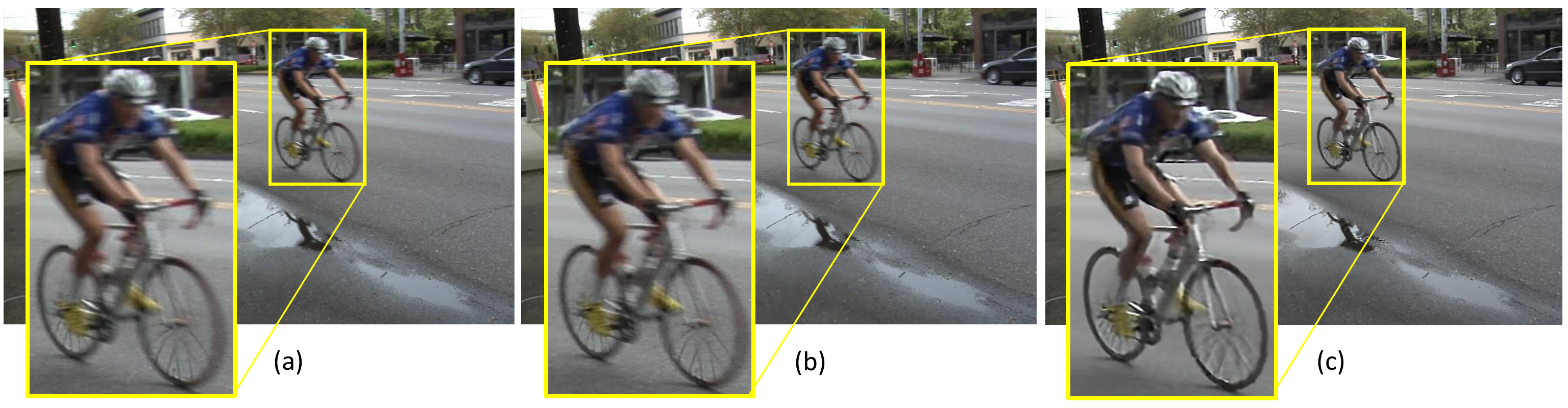}
	\end{center}
	\captionof{figure}{(a) Blurry frame from a dynamic scene. (b) Deblurring result by Cho et al.\cite{cho_siggraph2012}. (c) Our result.}
	\label{fig_teaser}
\end{figure*}

Finally, we provide a new realistic blur dataset with ground truth sharp frames captured by a high-speed camera to overcome the lack of realistic ground truth dataset in this field.
Though there have been some evaluation datasets for deblurring problem, they are not appropriate to carry out meaningful evaluation for the deblurring of spatially varying blurs.
First, synthetically generated uniform blur kernels and blurry images from sharp images were provided in the work of Levin et al.~\cite{Levin:2009}. 
Next, 6D camera motion in 3D space was recorded with a hardware-assisted camera to represent blur from camera shake during exposure time in the work of K{\"o}hler et al.~\cite{kohler2012recording}.
Moreover, there have been some recent approaches to generate synthetic dataset for the sake of machine learning algorithms. To benefit from large training data, lots of blur kernels and blurry images were synthetically generated. In the work of Xu et al.~\cite{xu2014deep}, more than
2500 blurry images are generated using decomposable symmetric kernels. Schuler et al.~\cite{Schuler_PAMI15} sampled naturally looking blur kernels with Gaussian Process, and Sun et al.~\cite{sun2015learning} used a set of linear kernels to synthesize blurry images. 
However, these datasets are generated under an assumption that the scene is static and cannot synthesize infinitely many blurs in real world. 
Real blurs in dynamic scenes are complex and spatially varying, 
so synthesizing realistic dataset is a difficult problem. 
To solve this problem, we construct a new blur dataset that provides pairs of realistically blurred videos and sharp videos with the use of a high-speed camera. 

Using the proposed dataset and real challenging videos as shown in Fig.\ref{fig_teaser},
we demonstrate the significant improvements of the proposed deblurring method in both quantitatively and qualitatively.
Moreover, we show empirically that more accurate optical flows are estimated by our method compared with the state-of-the-art optical flow method that can handle blurry images.

\section{More Generalized Video Deblurring}
Most conventional video deblurring methods suffer from the coexistence of various motion 
blurs from dynamic scenes because the motions cannot be fully parameterized using global or segment-wise blur models. 
To make things worse, frequent misfocus of moving objects in dynamic scenes yields more complex non-linear blurs 
combined with motion blurs.

To handle these joint motion and defocus blurs,
we propose a new blur model that estimates locally (pixel-wise) different blur kernels rather than global or segment-wise kernel estimation.
As blind deblurring problem is highly ill-posed,
we propose a single energy model consists of not only data and spatial regularization terms
but also a temporal term.
The model is expressed as follows:
\begin{equation}
\textbf{E} =  \textbf{E}_{data} + \textbf{E}_{temporal} + \textbf{E}_{spatial},
\label{equ_base}
\end{equation}
and the detailed models of each term in~(\ref{equ_base}) are given in the following sections.

\subsection{Data Model based on Kernel Approximation}
\begin{figure}[h]
	\begin{center}
		\includegraphics[width=\linewidth]{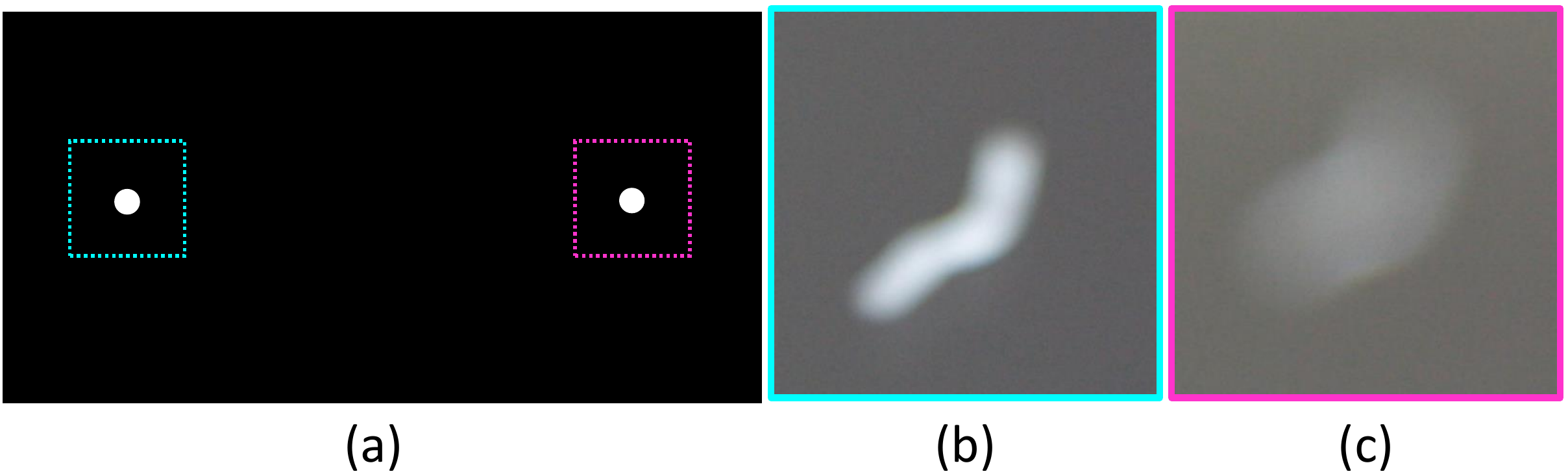}
	\end{center}
	\caption{(a) Two light sources. (b) Light streak of the focused light source. (c) Light streak of the defocused light source.}
	\label{fig_light_source}
\end{figure}

Motion blurs are generally caused by camera shake and moving objects,
and defocus blurs are mainly due to the aperture size, focal length, and the distance between camera and focused object.
These two different blurs are combined and yield more complex blurs in real video.
For example, 
Fig.~\ref{fig_light_source} shows how different the blurred images are when point light sources are captured by the same moving camera with and without defocused blur.
We observe that the light streak of the defocused light source is much smoother and non-linear in comparison with the focused one. Notably, the light streaks indicate the blur kernel shapes.

\begin{figure}[b]
	\begin{center}
		\includegraphics[width=\linewidth]{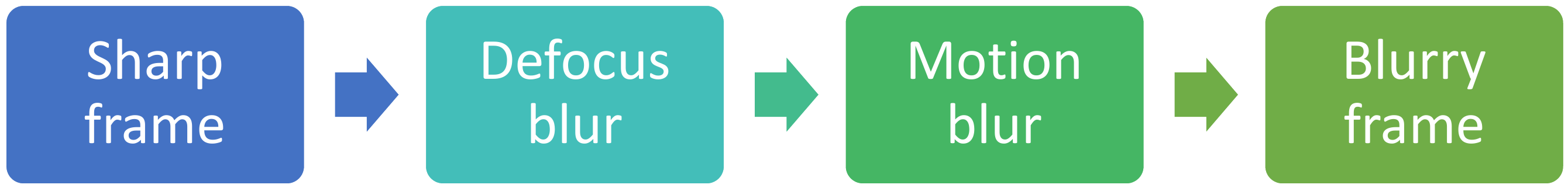}
	\end{center}
	\caption{Blurring process underlying in the proposed method.}
	\label{fig_blur_process}
\end{figure}

However, it is difficult to directly remove the complex blur in Fig.~\ref{fig_light_source} (c).
Thus, to alleviate the problem, we assume that the combined blur kernel can be decomposed into two different kernels, which are motion blur kernel and defocus blur kernel.
Our assumption holds when the depth change in the scene during exposure period is relatively small,
and it is acceptable since we treat video of rather short exposure time.
So, the underlying blurring-procedure can be modeled as sequential process of defocus blurring followed by motion blur as illustrated in Fig.~\ref{fig_blur_process}.

Note that in conventional video deblurring works~\cite{bar2007variational,cho_siggraph2012,li2010generating,Wulff:ECCV:2014}, 
the motion blurs of each frame are usually approximated by 
parametric models such as homography and affine models.
However, these kernel approximations are only valid when the motions are parameterizable within an entire frame or a segment, and cannot cope with spatially varying motion blurs.
To solve this problem, we approximate the pixel-wise motion blur kernel using bidirectional optical flows as suggested in previous works~\cite{Dai:2008,thkim_cvpr2014,Portz:2012}.

\begin{figure}[t]
	\begin{center}
		\includegraphics[width=\linewidth]{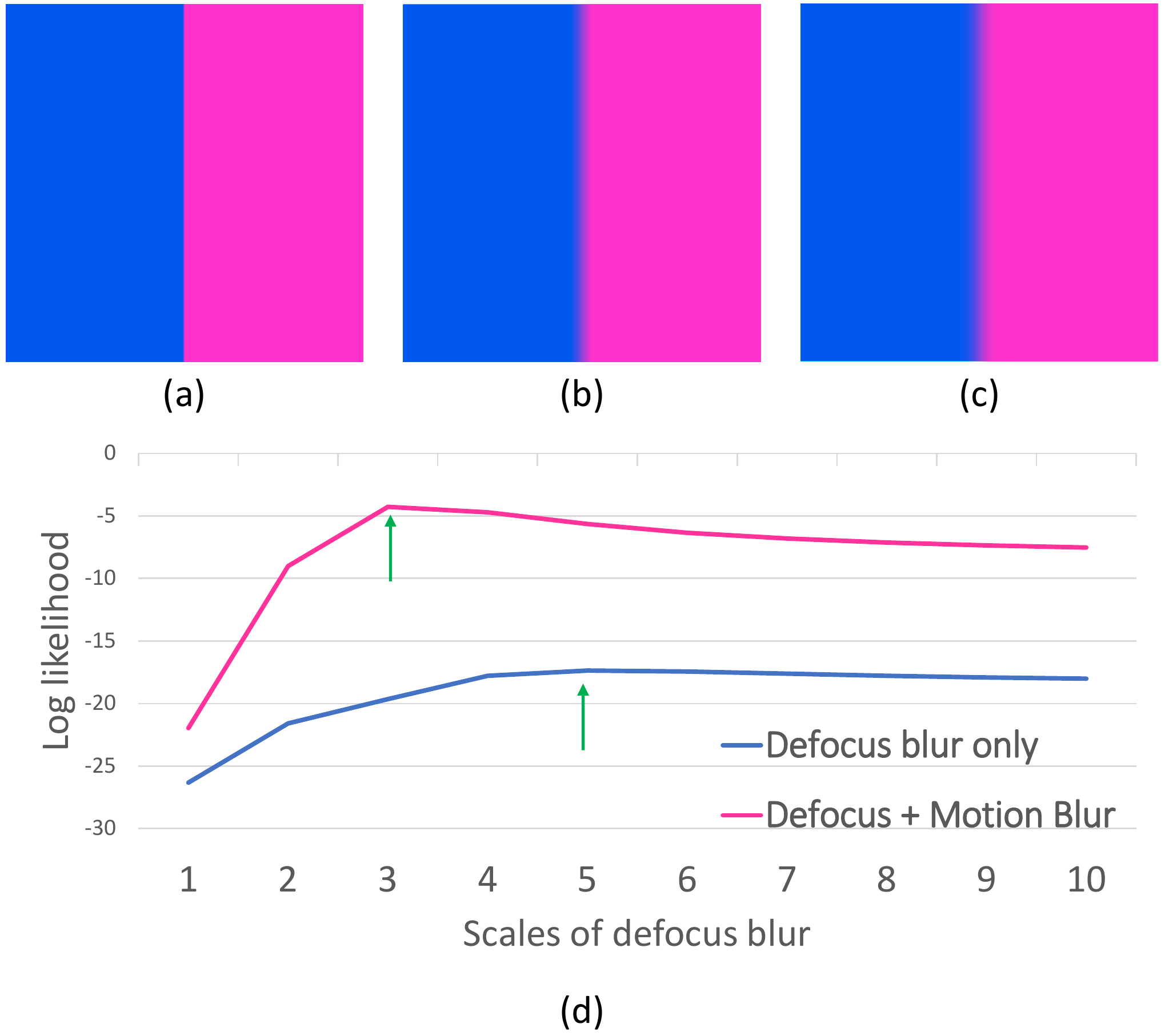}
	\end{center}
	\caption{(a) A sharp patch. (b) A patch blurred by defocus blur only (Gaussian blur with standard deviation 5). (c) A patch blurred by defocus blur (Gaussian blur with standard deviation 5) and motion blur (linear kernel with length 11). (d) Comparisons of fidelities at the center of the blurry patches by changing the scale of defocus blur. The ground truth scale of the defocus blur is 5 and the arrows indicate peaks estimated by ML estimator.}
	\label{fig_defocus_blur}
\end{figure}

Spatially varying defocus blur is approximated by using Gaussian or disc model in conventional works~\cite{kee2011modeling, zhu2013estimating}. Therefore, the defocus maps and scales of local blurs are determined by simply estimating the standard deviations of Gaussian models or the radii of disc models.
In particular, local image statistics is widely used to estimate spatially varying defocus blur.
Specifically, within a uniformly blurred patch, local frequency spectrum provides information on the blur kernel and can be used to determine the likelihood of specific blur kernel~\cite{zhu2013estimating}; thus scales of defocus blurs can be estimated by measuring the fidelities of the likelihood model.
However, it is difficult to apply this statistics-based technique when the blurry image has motion blurs in addition to defocus blurs. 
In Fig.~\ref{fig_defocus_blur}, we observe that the maximum likelihood (ML) estimator used in~\cite{zhu2013estimating} finds the optimal defocus blur kernel when a patch is blurred by only defocus blur,
however ML cannot estimate true defocus kernel when a blurry patch contains motion blur as well as defocus blur.
Therefore we cannot adopt local image statistics to remove defocus blurs in dynamic scenes with motion blurs. 
In this study, we approximate the pixel-wise varying defocus blur using Gaussian model, and determine the standard deviation of defocus blur by jointly estimating the defocus blur maps and latent frames unlike conventional works that utilize only local image statistics.

\begin{figure}[t]
	\begin{center}
		\includegraphics[width=\linewidth]{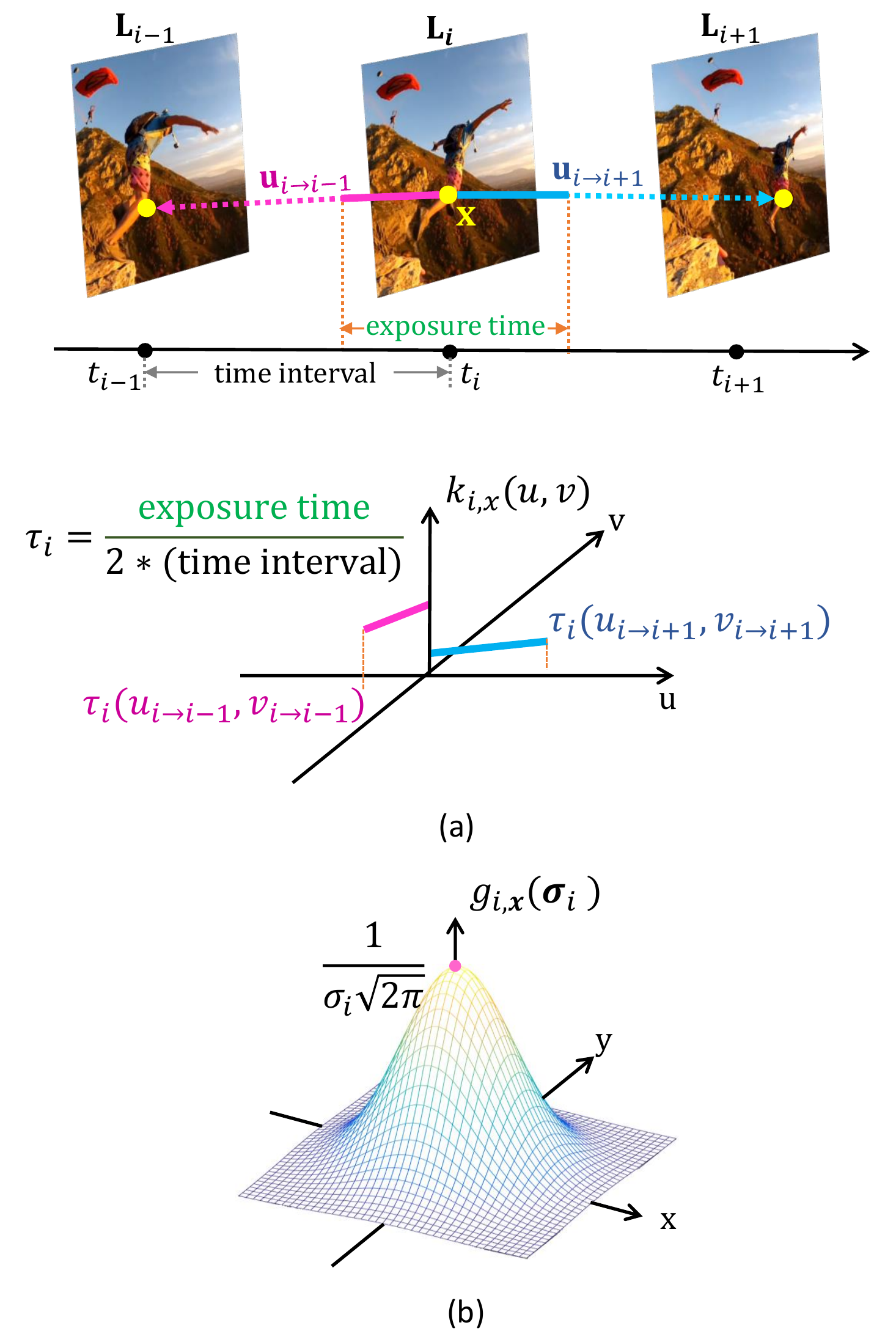}
	\end{center}
	\caption{(a) Bidirectional optical flows and corresponding piece-wise linear motion blur kernel at pixel location $\textbf{x}$. (b) Gaussian blur kernel with standard deviation $\sigma$ at $\textbf{x}$ to handle blur from defocus.}
	
	\label{fig_kernel}	
\end{figure}

Therefore, under two assumptions that the latent frames are blurred by defocus blurs, and subsequently blurred by motion, and the velocity of the motion is constant between adjacent frames, our blur model is expressed as follows:
\begin{equation}
\begin{split}
\textbf{B}_i(\textbf{x}) = (k_{i,\textbf{x}} \otimes g_{i,\textbf{x}}\otimes\textbf{L}_i)(\textbf{x})
\label{equ_blur_constraint},
\end{split}
\end{equation}
where $\textbf{B}_i$ and $\textbf{L}_i$ denote the blurry frame and the latent frame at the $i^{th}$ frame, respectively, and $\textbf{x}$ denotes pixel location on 2D image domain.
At $\textbf{x}$, the motion blur kernel is denoted as $k_{i,\textbf{x}}$
and the Gaussian blur kernel by defocus is denoted as $g_{i,\textbf{x}}$,
and the operator $\otimes$ means convolution.

To handle locally varying motion blurs, we should reduce the size of solution space using approximation and use parametrized kernel, because the solution space of locally varying kernel in video is extremely large; the dimension of kernel is $W\times H \times T \times w \times h$ when the size of image is $W\times H$, length of the image sequence is $T$, and the size of local kernel is $w \times h$.
Therefore, we approximate the motion blur kernel as piece-wise linear using bidirectional optical flows
as illustrated in Fig.~\ref{fig_kernel} (a).
Although our motion blur kernel is based on simple approximation, 
our model is valid since we assume that the videos have relatively short exposure time.
The pixel-wise kernel $k_{i,\textbf{x}}$ using bidirectional flow can be written by,

\begin{equation}
\begin{split}
&{k}_{i,\textbf{x}}(\textbf{u}_{i \rightarrow i+1}, \textbf{u}_{i \rightarrow i-1}) = \\
&\begin{cases}
\frac{\delta(u v_{i \rightarrow i+1} - v u_{i \rightarrow i+1})}{2\tau_i \|\textbf{u}_{i \rightarrow i+1}\|},&$if$~ u \in [0,\tau_i u_{i \rightarrow i+1}], v \in [0, \tau_i v_{i \rightarrow i+1}]\\
\frac{\delta(u v_{i \rightarrow i-1} - v u_{i \rightarrow i-1})}{2\tau_i \|\textbf{u}_{i \rightarrow i-1}\|},&$if$~ u \in (0,\tau_i u_{i \rightarrow i-1}], v \in (0, \tau_i v_{i \rightarrow i-1}]\\
0, &$otherwise$
\end{cases}
\end{split},
\label{equ_motion_blur_kernel_def}
\end{equation}
where $\textbf{u}_{i \rightarrow i+1}=(u_{i \rightarrow i+1}, v_{i \rightarrow i+1})$,
and $\textbf{u}_{i \rightarrow i-1}=(u_{i \rightarrow i-1}, v_{i \rightarrow i-1})$
denote pixel-wise bidirectional optical flows at frame $i$.
Camera duty cycle of the frame is $\tau_i$ and denotes relative exposure time as used in~\cite{li2010generating}, and $\delta$ denotes Kronecker delta.

\begin{figure}[t]
\begin{center}
\includegraphics[width=\linewidth]{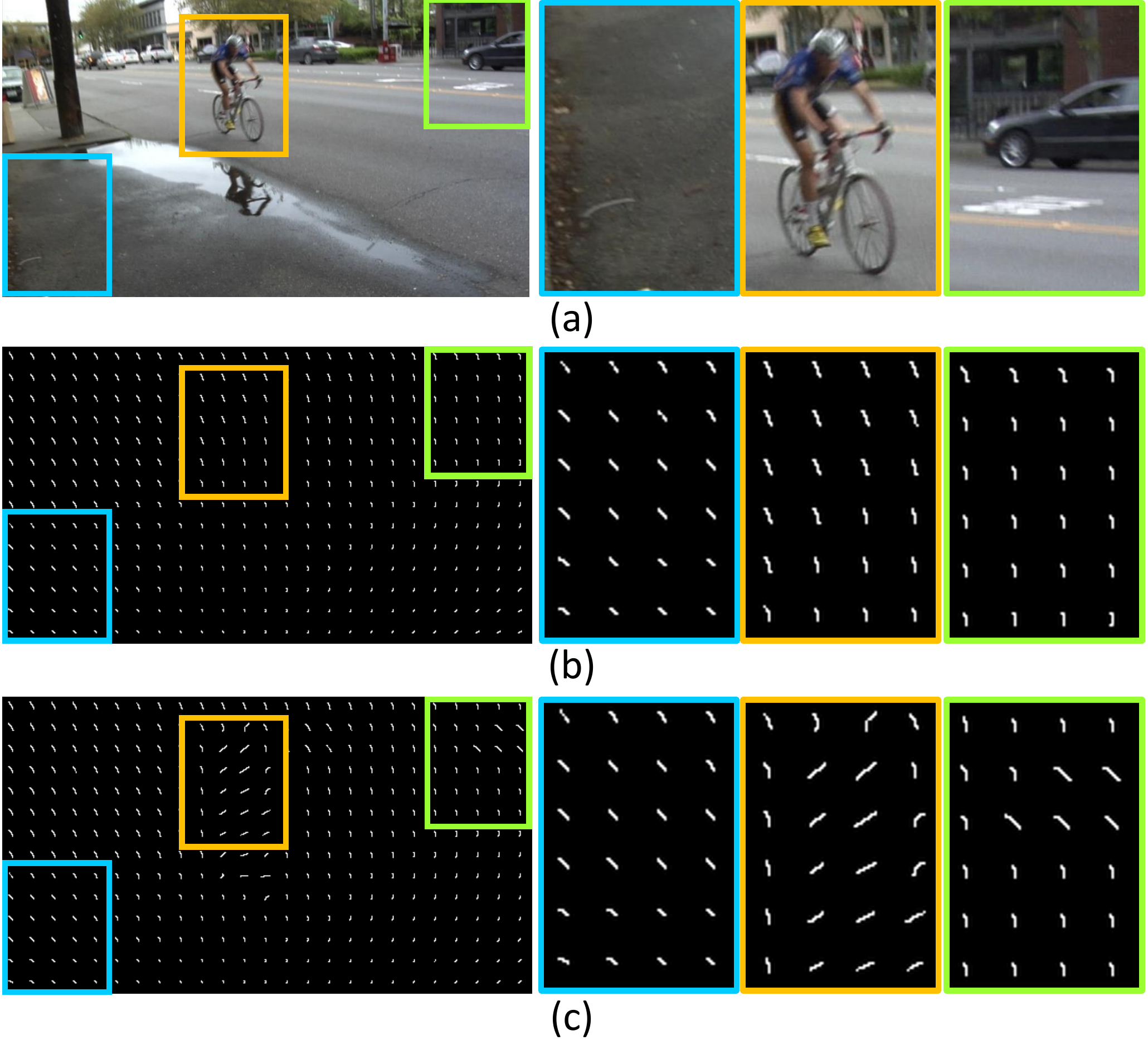}
\end{center}
\caption{(a) Blurry frame of a video in dynamic scene. (b) Locally varying kernel using homography. (c) Our pixel-wise varying motion blur kernel using bidirectional optical flows.}
\label{fig_kernel_difference}
\end{figure}

Using this pixel-wise motion blur kernel approximation,
we can easily manage multiple different motion blurs in a frame, unlike conventional methods.
The superiority of our locally varying kernel model is shown in Fig.~\ref{fig_kernel_difference}.
Our kernel model fits blurs from differently moving objects and camera shake much better
than the conventional homography-based model.

\begin{figure}[t]
	\begin{center}
		\includegraphics[width=\linewidth]{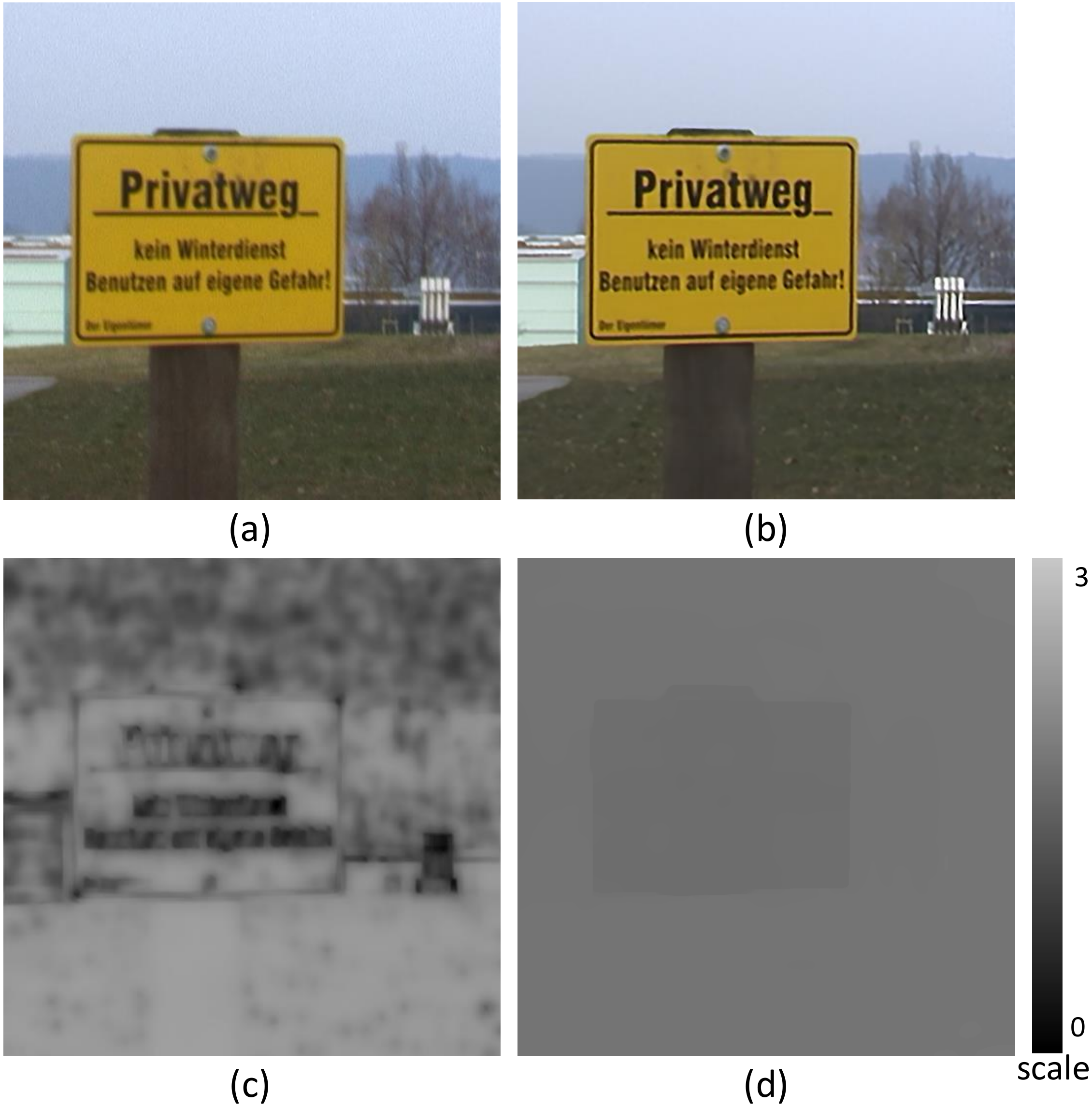}
	\end{center}
	\caption{(a) Real blurry frame. (b) Our jointly estimated latent frame. (c) Blur map of Shi et al.~\cite{shi2015just}. (d) Our defocus blur map. }	
	\label{fig_defocus_blur_map_comp_real}	
%

	\begin{center}		
	\includegraphics[width=\linewidth]{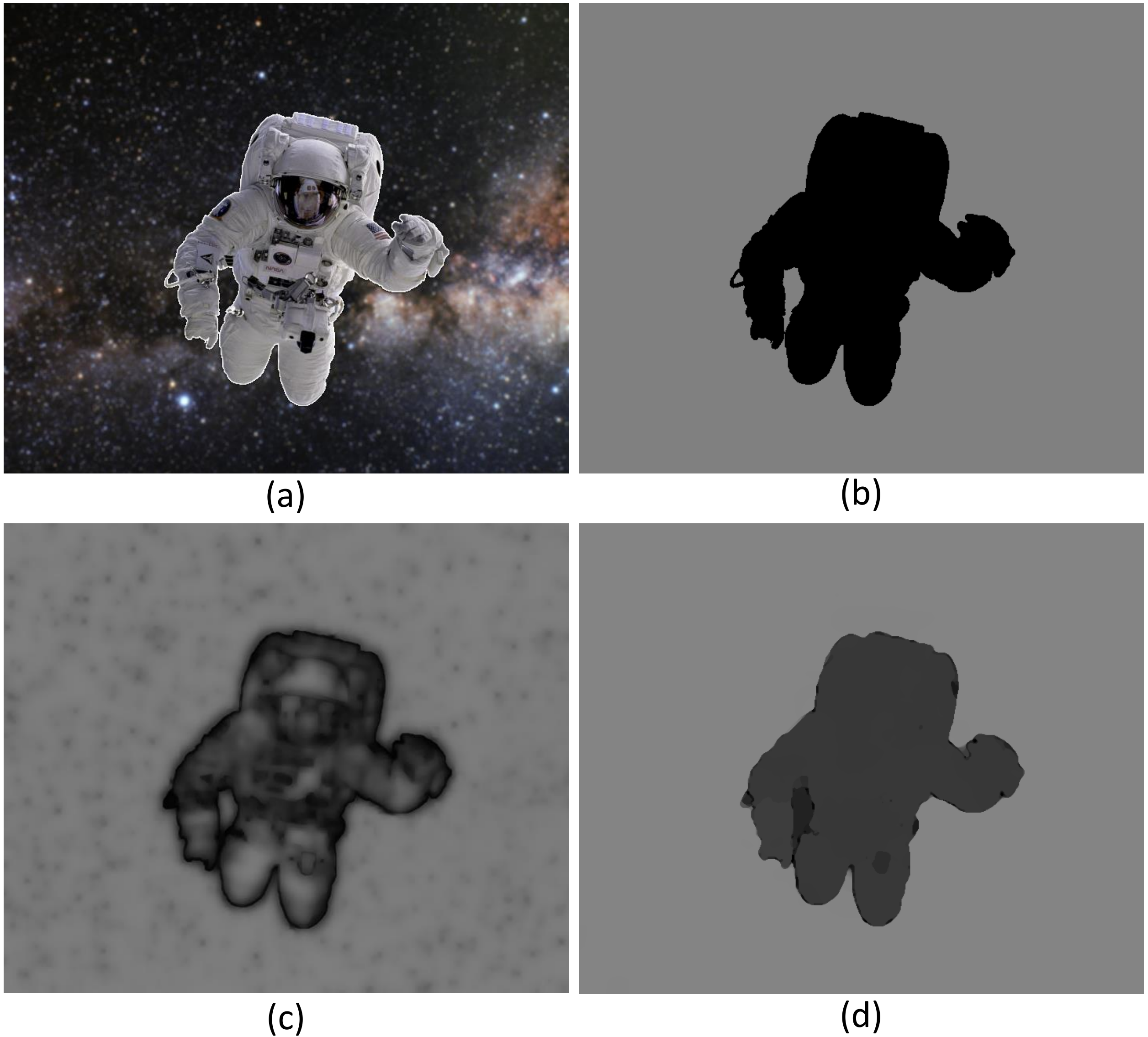}
	\end{center}
	
	\caption{(a) Partially blurred image which has sharp foreground and blurred background by Gaussian blur. (b) Ground truth blur map. (c) Defocus blur map of Shi et al.~\cite{shi2015just}. (d) Our defocus blur map.}
	\label{fig_defocus_blur_map_comp_synthetic}

\end{figure}

Moreover, we approximate the spatially varying defocus blur kernel using Gaussian model
as shown in Fig.~\ref{fig_kernel} (b), and estimate the pixel-wise different standard deviation $\sigma_i$ of the Gaussian kernel $g_{i,\textbf{x}}$.
Although we cannot utilize the features in the blurred frame,
which has been used significantly in conventional methods~\cite{zhu2013estimating,shi2015just},
due to combined motion blurs, we determine the scales of defocus blurs with the simultaneous estimation of latent frames and achieve significant improvements in comparison with the state-of-the-art defocus blur map estimator~\cite{shi2015just} when there exist both motion and defocus blurs in a real blurry frame as shown in Fig.~\ref{fig_defocus_blur_map_comp_real}. Moreover, even when the motion blurs are not existing, we achieve competitive result as shown in Fig.~\ref{fig_defocus_blur_map_comp_synthetic}.

Now, the proposed data model that handles both motion and defocus blurs
is expressed as follows:
%
\begin{equation}
\begin{split}
&\textbf{E}_{data}({\textbf{L}, \textbf{u}, \bm{\sigma}, \textbf{B}}) =\\ &\lambda\sum_{i} \sum_{\partial_*}  \|\partial_*  \textbf{K}_i(\tau_i, \textbf{u}_{i \rightarrow i+1}, \textbf{u}_{i \rightarrow i-1}) \textbf{G}_i(\bm{\sigma}_i)\textbf{L}_i -\partial_* \textbf{B}_i \|^2,
\end{split}
\label{equ_data}
\end{equation}
where the row vector of the motion blur kernel matrix $\textbf{K}_i$, which corresponds to the motion blur kernel at pixel $\textbf{x}$, 
is the vector form of $k_{i,\textbf{x}}(.)$, and its elements are non-negative and their sum is equal to one. 
Similarly, the row vector of the defocus blur kernel matrix $\textbf{G}_i$, which corresponds to the Gaussian kernel at $\textbf{x}$, is the vector form of $g_{i,\textbf{x}}(.)$ and $\bm{\sigma}_i$ denotes the scales (standard deviation of Gaussian kernel) of defocus blurs.
Linear operator $\partial_*$ denotes the Toeplitz matrices corresponding to the partial 
(e.g., horizontal and vertical) derivative filters.
Parameter $\lambda$ controls the weight of the data term,
and $\textbf{L}$, $\textbf{u}$, $\bm{\sigma}$, and $\textbf{B}$ denote the set of latent frames,
optical flows, scales of defocus blurs and blurry frames, respectively.

\subsection{A new Optical Flow Constraint and Temporal Regularization}
As discussed above, 
to remove locally varying motion blurs, we employ bidirectional optical flow model
in~(\ref{equ_data}).
However, for optical flow estimation, conventional optical flow constraints such as brightness constancy and gradient constancy can not be utilized directly, since such constraints do not hold between two blurry frames.
A blur-aware optical flow estimation method from blurry images has been proposed by Portz et al.~\cite{Portz:2012}. This method is based on the commutative law of shift-invariant kernels such that the brightness of the corresponding points is constant after convolving the blur kernel of each image with the other image. However, the commutative law does not hold when the motion is not translational and when the blur kernels vary spatially. Therefore, this approach only works when the motion is very smooth.

To address this problem, we propose a new model that estimates optical flow between two latent sharp frames to enable abrupt changes in motions and the blur kernels.
In using this model, we need not restrict our motion blur kernels to be shift invariant.
Our model is based on the conventional optical flow constraint between latent frames,
that is, brightness constancy.
The formulation of this model is given by,
\begin{equation}
\begin{split}
\textbf{E}_{temporal}&(\textbf{L},\textbf{u}) = \\
&\mu  \sum_{i} \sum_{\textbf{x}} \sum_{n = -N}^{N}  | \textbf{L}_i(\textbf{x}) - \textbf{L}_{i + n}(\textbf{x} + \textbf{u}_{i \rightarrow i+n})  |,
\end{split}
\label{equ_temporal}
\end{equation}
where $n$ denotes the index of neighboring frames at $i$, 
and the parameter $\mu$ controls the weight.
We apply the robust $L_1 $ norm for robustness against outliers and occlusions.

\begin{figure}[t]
	\begin{center}
		\includegraphics[width=\linewidth]{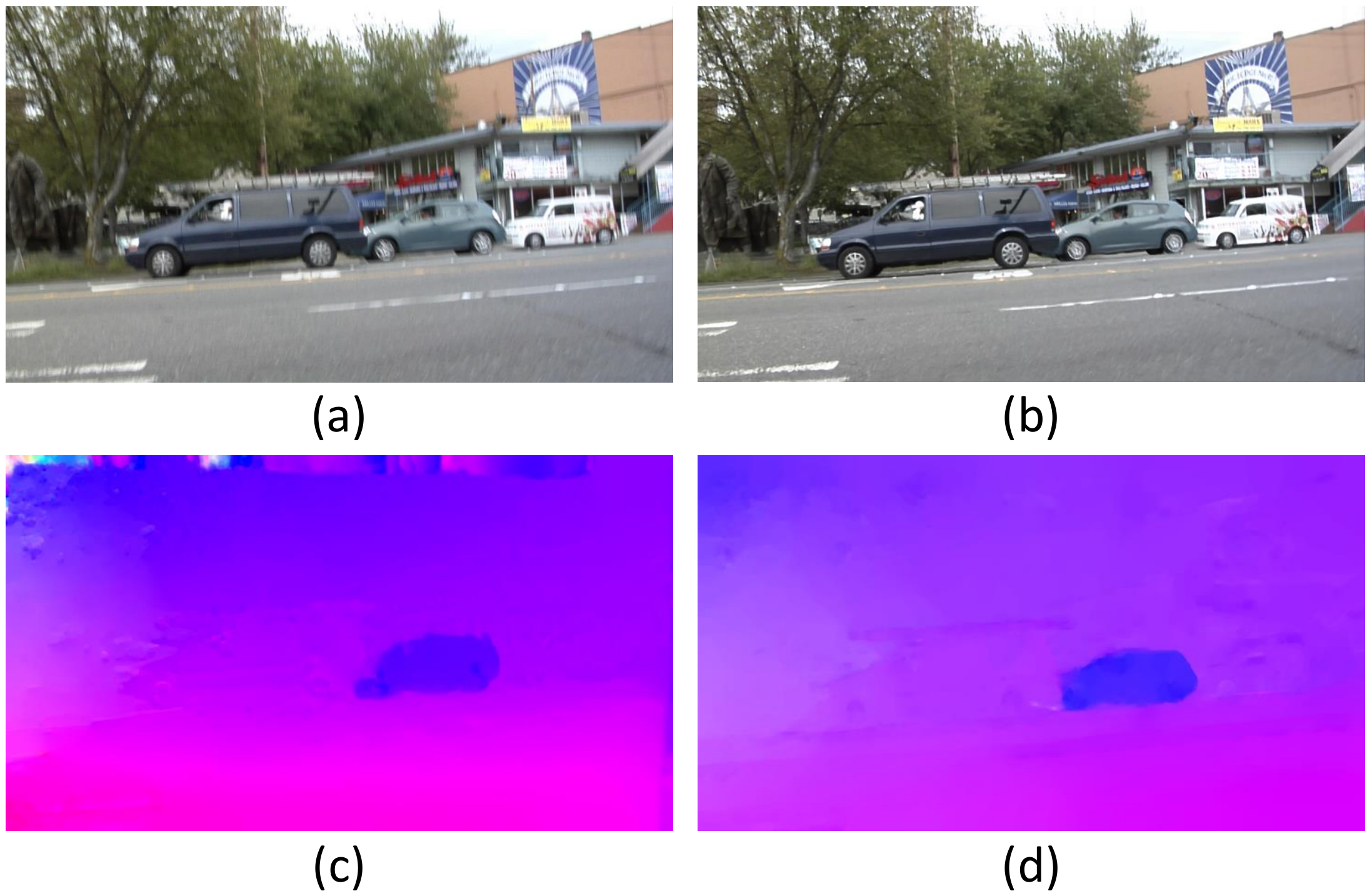}
	\end{center}
	\caption{(a) Real blurry frame. (b) Our jointly estimated latent frame. (c) Color coded optical flow from~\cite{Portz:2012}. (d) Our optical flow.}
	\label{fig_flow}
\end{figure}

Notably, a major difference between the proposed model and the conventional optical flow estimation methods is that our problem is a joint problem.
That is, the latent frames and optical flows should be solved simultaneously.
Therefore, the proposed model in~(\ref{equ_temporal}) estimates the latent frames which are temporally coherent among neighboring frames, and optical flows between neighboring frames, simultaneously.
Therefore we can estimate accurate flows at the motion boundaries as shown in Fig.~\ref{fig_flow}.
Notice that, our flows at the motion boundaries of moving car is much clearer in comparison with the blur-aware flow estimation method by~\cite{Portz:2012}.

\subsection{Spatial Regularization}
To alleviate the difficulties of highly ill-posed deblurring, optical flow estimation, 
and defocus blur map estimation problems, it is important to adopt spatial regularizers.
In doing so, we enforce spatial coherence to penalize spatial fluctuations
while allowing discontinuities in latent frames, flow fields, and defocus blur maps.
With the assumption that spatial priors for the latent frames, optical flows, and defocus blur maps are independent, we can formulate the spatial regularization as follows:
\begin{equation}
\begin{split}
\textbf{E}_{spatial}(\textbf{L},\textbf{u}) = &\sum_i  | \nabla \textbf{L}_i | + \nu_{\sigma} \sum_i\sum_{\textbf{x}}g_i(\textbf{x})|\nabla \bm{\sigma_i}| + \\ 
& \nu_{u}\sum_i\sum_{\textbf{x}}\sum_{n=-N}^{N}  g_i(\textbf{x})| \nabla \textbf{u}_{i \rightarrow i+n} |),
\end{split}
\label{equ_spatial}
\end{equation}
wherer parameters $\nu_{\sigma}$ and $\nu_{u}$ control the weights of the second and third terms.

The first term in (\ref{equ_spatial}) denotes the spatial regularization term for the latent frames. Although more sparse $L_p$ norms (e.g. $p = 0.8$) fit the gradient statistics of natural sharp images better~\cite{Krishnan:2009,Krishnan:2011,Levin:PAMI2007}, we use conventional total variation (TV) based regularization~\cite{hu_cvpr2014_depthdeblur,thkim:2013,thkim_cvpr2014},
as TV is computationally less expensive and easy to minimize. 
The second and third terms enforce spatial smoothness for defocus blur maps and optical flows, respectively.
These regularizers are also based on TV regularization,
and coupled with edge-map to preserve discontinuities at the edges in both vector fields.
Similar to the edge-map used in conventional optical flow estimation method~\cite{thkim_iccv2013optical}, 
our edge-map is expressed as follows:
\begin{equation}
g_i(\textbf{x}) =  \exp(- (\frac {|\nabla {\textbf{L}}_{i,0}|^2}{v_I})),
\end{equation}
where the fixed parameter $v_I$ controls the weight of the edge-map,
and ${\textbf{L}}_{i,0}$ is an initial latent image in the iterative optimization framework.

\section{Optimization Framework}
Under the condition that the camera duty cycle $\tau_i$ is known, by combining  $\textbf{E}_{data}$, $\textbf{E}_{temporal}$, and $\textbf{E}_{spatial}$, 
we can have the final objective function as follows:
and the final objective function when camera duty cycle $\tau_i$ is known becomes as follows:
\begin{equation}
\begin{split}
\min_{\textbf{L},\textbf{u},\bm{\sigma}} ~ \lambda & \sum_{i} \sum_{\partial_*}  \|\partial_*  \textbf{K}_i(\textbf{u}_{i \rightarrow i+1}, \textbf{u}_{i \rightarrow i-1}) \textbf{G}_i(\bm{\sigma_i})\textbf{L}_i -\partial_* \textbf{B}_i \|^2 +\\
&\mu \sum_{i} \sum_{\textbf{x}} \sum_{n = -N}^{N}  | \textbf{L}_i(\textbf{x}) - \textbf{L}_{i + n}(\textbf{x} + \textbf{u}_{i \rightarrow i+n})  | + \\
&\sum_i  | \nabla \textbf{L}_i | + \nu_{\sigma} \sum_i\sum_{\textbf{x}}g_i(\textbf{x})|\nabla \bm{\sigma_i}| + \\ 
& \nu_{u}\sum_i\sum_{\textbf{x}}\sum_{n=-N}^{N}  g_i(\textbf{x})| \nabla \textbf{u}_{i \rightarrow i+n} |).
\end{split}
\label{equ_final}
\end{equation}
Note that contrast with Cho et al.~\cite{cho_siggraph2012} that performs multiple approaches sequentially,
our model finds a solution by minimizing the proposed single objective function in~(\ref{equ_final}).
However, because of its non-convexity, our model needs to adopt a practical optimization method to obtain an approximated solution.
Therefore, we divide the original problem into several simple sub-problems and then use conventional iterative 
and alternating optimization techniques~\cite{Cho:2009,thkim_cvpr2014, Wulff:ECCV:2014} to minimize the original non-convex objective function.
In the following sections, 
we introduce efficient solvers and describe how to estimate unknowns $\textbf{L}$, $\textbf{u}$ and $\bm{\sigma}$ alternatively.

\subsection{Sharp Video Restoration}
If the motion blur kernels $\textbf{K}$ and the defocus blur kernels $\textbf{G}$ are fixed, 
then the objective function in~(\ref{equ_final}) becomes convex with respect to $\textbf{L}$, and it can be reformulated as follows:
\begin{equation}
\begin{split}
\min_{\textbf{L}} ~~\lambda  &\sum_i \sum_{\partial_*} \|\partial_*  \textbf{K}_i \textbf{G}_i \textbf{L}_i -\partial_* \textbf{B}_i \|^2 + \sum_i| \nabla \textbf{L}_i | +\\
&\mu  \sum_i  \sum_{\textbf{x}} \sum_{n = -N}^{N} | \textbf{L}_i(\textbf{x}) - \textbf{L}_{i+n}(\textbf{x}+\textbf{u}_{i \rightarrow i+n}) |.
\end{split} 
\label{equ_video_restoration}
\end{equation}
To restore the latent frames $\textbf{L}$, 
we adopt the conventional convex optimization method proposed in~\cite{Chambolle:2011}, 
and derive the primal-dual update scheme as follows:
\begin{equation}\begin{cases}
\textbf{s}^{m+1} = \frac {\textbf{s}^m + \eta_L  \textbf{A} \textbf{L}^m}{\max(\textbf{1}, ~\text{abs}(\textbf{s}^m + \eta_L \textbf{A} \textbf{L}^m ) )}  \\
\\
\textbf{q}^{m+1} = \frac {\textbf{q}^m + \eta_L \mu \textbf{D} \textbf{L}^m}
{\max(\textbf{1}, ~\text{abs}(\textbf{q}^m + \eta_L
\mu \textbf{D} \textbf{L}^m))}  \\
\\
\begin{split}
&\textbf{L}^{m+1} = \arg \min_{\textbf{L}^{m+1}}  \lambda \sum_i\sum_{\partial_*}  \| \partial_* \textbf{K}_i \textbf{G}_i \textbf{L}_i^{m+1} - \partial_*\textbf{B}_i\|^2 +\\
\\
& {\frac{\| \textbf{L}^{m+1} - (\textbf{L}^m - \epsilon_L( \textbf{A}^T\textbf{s}^{m+1} + \mu{\textbf{D}}^T \textbf{q}^{m+1} ))\| ^2}{2\epsilon_L}},
\end{split}
\end{cases}
\label{equ_update_L}
\end{equation}
where $m\geq0$ indicates the iteration number,
and, $\textbf{s}^m$ and $\textbf{q}^m$ denote the dual variables of the concatenated latent frames $\textbf{L}^m$.
Parameters $\eta_L$ and $\epsilon_L$ denote the update steps.
Linear operator $\textbf{A}$ calculates the spatial difference between neighboring pixels,
and operator $\textbf{D}$ calculates the temporal differences among neighboring frames using fixed optical flows.
The last formulation in~(\ref{equ_update_L}) is to update and optimize the primal variable $\textbf{L}^{m+1}$,
and we apply the conjugate gradient method to minimize it since it is a quadratic function.

\subsection{Optical Flows Estimation}
Note that , although the latent frames $\textbf{L}$ and the defocus blur kernels $\textbf{G}$ are fixed, the temporal coherence term $\textbf{E}_{temporal}$ and the data term $\textbf{E}_{data}$ are still non-convex.
So, let us denote those two terms as a non-convex function $\rho_{u}(.)$ as follows:
\begin{equation}
\begin{split}
\rho_{u}(\textbf{u})= 
\mu \sum_i \sum_{\textbf{x}}\sum_{n=-N}^N  | \textbf{L}_i(\textbf{x}) - \textbf{L}_{i+n}(\textbf{x}+\textbf{u}_{i \rightarrow i+n}) |+\\
\lambda \sum_i \sum_{\partial_*}  \| \partial_* \textbf{K}_i(\textbf{u}_{i \rightarrow i+1}, \textbf{u}_{i \rightarrow i-1}) \textbf{G}_i\textbf{L}_i - \partial_* \textbf{B}_i\|^2.
\end{split}
\end{equation}
To find the optimal optical flows $\textbf{u}$, 
we first convexify the non-convex function $\rho_u(.)$ by applying the first-order Taylor expansion.
Similar to the technique in~\cite{thkim_cvpr2014}, we linearize the function near an initial $\textbf{u}_0$ in the iterative process as follows:
\begin{equation}
\rho_u(\textbf{u}) \approx \rho_u(\textbf{u}_0) + \nabla \rho_u(\textbf{u}_0)^T(\textbf{u}-\textbf{u}_0).
\end{equation}
In doing so, (\ref{equ_final}) can be approximated by a convex function w.r.t $\textbf{u}$ as follows:
\begin{equation}
\begin{split}
\min_{\textbf{u}} \rho_u(\textbf{u}_0) +  
&\nabla \rho_u(\textbf{u}_0)^T(\textbf{u}-\textbf{u}_0) + \\
&\nu_{u}\sum_i \sum_{\textbf{x}}\sum_{n=-N}^{N} g_i(\textbf{x}) |\nabla \textbf{u}_{i\rightarrow i+n}|.
\end{split}
\label{equ_convex_data}
\end{equation}

Now, we can apply the convex optimization technique in~\cite{Chambolle:2011}
to the approximated convex function,
and the primal-dual update process is expressed as follows:
\begin{equation}\begin{cases}
\textbf{p}^{m+1} = \frac {\textbf{p}^m + \eta_u (\nu_u \textbf{W}_u \textbf{A}_u) \textbf{u}^m}{\max(\textbf{1},  ~\text{abs}(\textbf{p}^m + \eta_u (\nu_u \textbf{W}_u\textbf{A}_u) \textbf{u}^m))}  \\
\\
\textbf{u}^{m+1} = (\textbf{u}^m - \epsilon_u(\nu_u \textbf{W}_u\textbf{A}_u)^T \textbf{p}^{m+1}) - \epsilon_u \nabla\rho_u(\textbf{u}_0),
\end{cases}
\label{equ_update_motion}
\end{equation}
where $\textbf{p}$ denotes the dual variable of $\textbf{u}$ on the vector space.
Weighting matrix $\textbf{W}_u$ is diagonal, and its sub-matrix associated with $\textbf{u}_{i \rightarrow i+n}$ is defined as $\text{diag}(g_i(\textbf{x}))$.
Linear operator $\textbf{A}_u$ calculates the spatial difference between four nearest neighboring pixels,
and parameters $\eta_u$ and $\epsilon_u$ denote the update steps.

\subsection{Defocus Blur Map Estimation}
When the latent frames $\textbf{L}$ and the motion blur kernels $\textbf{K}$ are fixed,
we can estimate the defocus blur maps from~(\ref{equ_final}). 
However, the data term is non-convex, and thus an approximation technique is required to optimize the objective function.
Similar to our optical flows estimation technique, we approximate and convexify the original function using linearization. 

First, we define a non-convex data function $\rho_{\bm{\sigma}}(.)$, and approximate it near an initial values $\bm{\sigma}_{0}$ as follows:
\begin{equation}
\begin{split}
\rho_{\sigma}(\bm{\sigma}) &= \lambda \sum_i\sum_{\partial_*}  \| \partial_* \textbf{K}_i \textbf{G}_i(\bm{\sigma}_i) \textbf{L}_i - \partial_* \textbf{B}_i \|^2\\
&\approx \rho_{\sigma}(\bm{\sigma}_0) + \nabla \rho_{\sigma}(\bm{\sigma}_0)^T(\bm{\sigma}-\bm{\sigma}_0),
\end{split}
\end{equation}
and the approximated convex function for defocus blur map estimation is given by,
\begin{equation}
\begin{split}
\min_{\bm{\sigma}}~  \rho_{\sigma}(\bm{\sigma}_{0}) +  
&\nabla \rho_{\sigma}(\bm{\sigma}_0)^T(\bm{\sigma}-\bm{\sigma}_0) + \\ &\nu_{\sigma}\sum_i\sum_{\textbf{x}}g_i(\textbf{x}) |\nabla \bm{\sigma}_i|.
\end{split}
\label{equ_convex_data_defocus}
\end{equation}

Similarly, (\ref{equ_convex_data_defocus}) can be optimized using \cite{Chambolle:2011},
and the primal-dual update formulation is given by,
\begin{equation}\begin{cases}
\textbf{r}^{m+1} = \frac {\textbf{r}^m + \eta_{\sigma} (\nu_{\sigma} \textbf{W}_{\sigma}\textbf{A}) \bm{\sigma}^m}{\max(\textbf{1},  ~\text{abs}(\textbf{r}^m + \eta_{\sigma} (\nu_{\sigma} \textbf{W}_{\sigma}\textbf{A}) \bm{\sigma}^m))}  \\
\\
\bm{\sigma}^{m+1} = (\bm{\sigma}^m - \epsilon_{\sigma}(\nu_{\sigma}\textbf{W}_{\sigma}\textbf{A})^T \textbf{r}^{m+1}) - \epsilon_{\sigma} \nabla \rho_{\sigma}(\bm{\sigma}_0),
\end{cases}
\label{equ_update_motion}
\end{equation}
where $\textbf{r}$ denotes the dual variable of $\bm{\sigma}$ on the vector field.
Weighting matrix $\textbf{W}_{\sigma}$ is diagonal, and its sub-matrix associated with $\bm{\sigma}_i$ is defined as $\text{diag}(g_i(\textbf{x}))$.
Parameters $\eta_{\sigma}$ and $\epsilon_{\sigma}$ denote the update steps.

\section{Implementation Details}
\begin{figure}[t]
\begin{center}
\includegraphics[width=\linewidth]{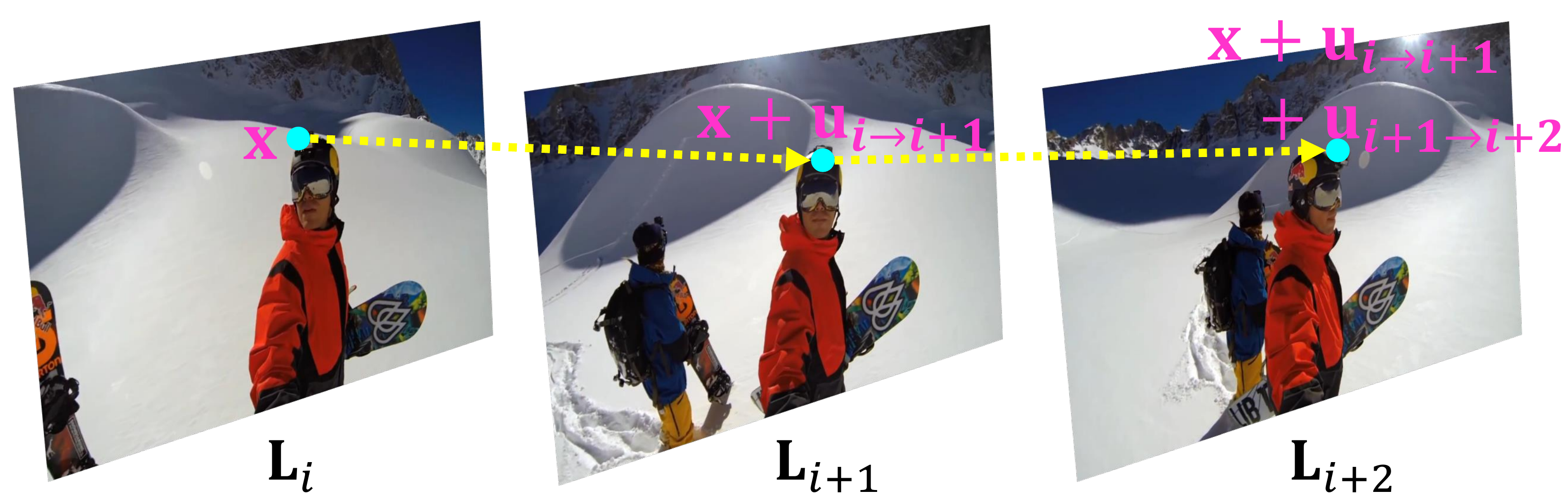}
\end{center}
\caption{Temporally consistent optical flows over three frames.}
\label{fig_corr}
\end{figure}

To handle large blurs and guide fast convergence, we implement our algorithm on the conventional coarse-to-fine framework with empirically determined parameters.
In the coarse-to-fine framework, we build image pyramid with 17 levels 
for a high-definition(1280x720) video, and use the scale factor 0.9.

Moreover, to reduce the number of unknowns in optical flows, 
we only estimate $\textbf{u}_{i \rightarrow i+1 }$ and $\textbf{u}_{i \rightarrow i-1 }$.
We approximate $\textbf{u}_{i \rightarrow i+2 }$ using $\textbf{u}_{i \rightarrow i+1 }$ and $\textbf{u}_{i+1 \rightarrow i+2 }$. 
For example, it satisfies, $\textbf{u}_{i \rightarrow i+2} = \textbf{u}_{i \rightarrow i+1} + \textbf{u}_{i+1 \rightarrow i+2}$, as illustrated in Fig.~\ref{fig_corr}, and we can easily apply this for $n \neq 1$.
Please see our publicly available source code for more details~\footnote{\url{http://cv.snu.ac.kr/research/~VD/}}.

The overall process of our algorithm is in Algorithm \ref{algorithm_overall}.
Further details on initialization, estimating the duty cycle $\tau_i$
and post-processing step that reduces artifacts are given below.

\begin{algorithm}[h]
\caption{Overview of the proposed method}
\begin{algorithmic}[1]
\renewcommand{\algorithmicrequire}{\textbf{Input:}}
\renewcommand{\algorithmicensure}{\textbf{Output:}}
\REQUIRE Blurry frames $\textbf{B}$
\ENSURE  Latent frames $\textbf{L}$, optical flows $\textbf{u}$, and defocus blur maps $\bm{\sigma}$
\STATE {Initialize $\textbf{u}$, ${\tau_i}$, and $\bm{\sigma}$. (Sec. 4.1)}
\STATE {Build image pyramid.}
\STATE {Restore $\textbf{L}$ with fixed $\textbf{u}$ and $\bm{\sigma}$. (Sec. 3.1)}
\STATE {Estimate $\textbf{u}$ with fixed $\textbf{L}$ and $\bm{\sigma}$. (Sec. 3.2)}
\STATE {Estimate $\bm{\sigma}$ with fixed $\textbf{L}$ and $\textbf{u}$. (Sec. 3.3)}
\STATE {Detect occlusion and perform post-processing. (Sec 4.2)}
\STATE {Propagate variables to the next pyramid level if exists.}
\STATE {Repeat steps 3-7 from coarse to fine pyramid level.}
\end{algorithmic} \label{algorithm_overall}	
\end{algorithm}

\begin{figure*}[t]
	\begin{center}
		\includegraphics[width=\linewidth]{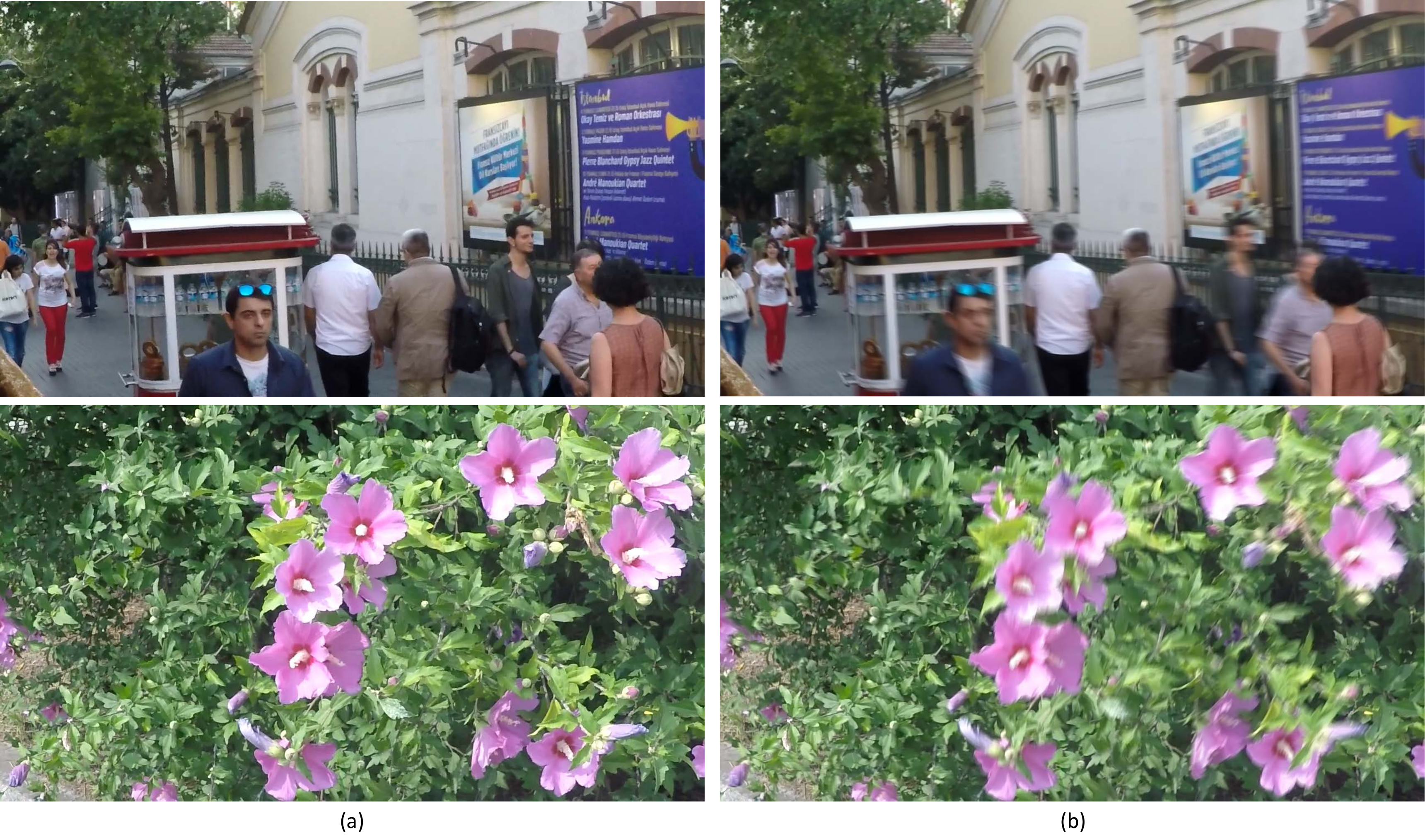}
	\end{center}
	\caption{(a) Ground truth sharp frames. (b) Generated blurry frames.
		Spatially varying blurs by object motions and camera shakes are synthesized realistically.}
	\label{fig_dataset_example}
\end{figure*}

\subsection{Initialization and Duty Cycle Estimation}
In this study, we assume that the camera duty cycle $\tau_i$ is known for every frame.
However, when we conduct deblurring with conventional datasets, which do not provide exposure information,
we apply the technique proposed in~\cite{cho_siggraph2012} to estimate the duty cycle.
Contrary to the original method~\cite{cho_siggraph2012},
we use optical flows instead of homographies to obtain initially approximated blur kernels.
Therefore, we first estimate flow fields from blurry images with~\cite{wedel2009improved}, which runs in near real-time.
We then use them as initial flows and approximate the kernels to estimate the duty cycle.
Moreover, we use $\bm{\sigma} = 0.8$ as initial defocus blur scale.

\subsection{Occlusion Detection and Refinement}
Our pixel-wise kernel estimation naturally results in approximation error
and it causes problems such as ringing artifacts.
Specifically, our data model in (\ref{equ_data}), 
and temporal coherence model in (\ref{equ_temporal}) are invalid at occluded regions.

To reduce such artifacts from kernel approximation errors and occlusions, we use spatio-temporal filtering as a post-processing:
\begin{equation}
\textbf{L}^{m+1}_i(\textbf{x})^ = \frac{1}{Z(\textbf{x})}\sum_{n=-N}^N \sum_\textbf{y} w_{i,n}(\textbf{x},\textbf{y})\cdot \textbf{L}_{i+n}^m(\textbf{y}),
\label{equ_spatio_temporal_filter}
\end{equation}
where \textbf{y} denotes a pixel in the 3x3 neighboring patch at location $(\textbf{x}+\textbf{u}_{i \rightarrow i+n})$
and $Z$ is the normalization factor (e.g. ${Z(\textbf{x})} = \sum_{n=-N}^N \sum_\textbf{y} w_{i,n}(\textbf{x},\textbf{y})$).
Notably, we enable $n=0$ in (\ref{equ_spatio_temporal_filter}) for spatial filtering.
Our occlusion-aware weight $w_{i,n}$ is defined as follows:
\begin{equation}
w_{i,n}(\textbf{x},\textbf{y}) = o_{i,n}(\textbf{x},\textbf{y}) \cdot \exp(- \frac{  \| P_i(\textbf{x}) - P_{i+n}(\textbf{y}) \|^2 } {2\sigma_w^2} ),
\end{equation}
where occlusion state $o_{i,n}(\textbf{x},\textbf{y}) \in \{0.01, 1\} $ is determined by cross-checking forward and backward flows similar to the occlusion detection technique used in~\cite{rhemann2011fast}.
The 5x5 patch $P_i(\textbf{x})$ is centered at $\textbf{x}$ in frame $i$.
The similarity control parameter $\sigma_w$ is fixed as $\sigma_w = 25/255$.

\section{Motion Blur Dataset}

Because conventional evaluation datasets for deblurring~\cite{Levin:2009, kohler2012recording}
are generated under static scene assumption, complex and spatially varying blurs in dynamic scenes are not provided. Therefore, in this section, we provide a new method generating blur dataset for the quantitative evaluation of non-uniform video deblurring algorithms and later studies of learning-based deblurring approaches.

\subsection{Dataset Generation}

As we assume motion blur kernels can be approximated by using bidirectional optical flows in~(\ref{equ_motion_blur_kernel_def}), we can generate blurry frames adversely by averaging consecutive frames whose relative motions between two neighboring frames are smaller than 1 pixel.
In doing so, we use GOPRO Hero4 hand-held camera which supports taking 240 fps video of 1280x720 resolution. 
Similar approach was introduced in ~\cite{agrawal2009optimal}, which uses high-speed camera to generate blurry images. However, they captured only linearly moving objects with a static camera.

Our captured videos include various dynamic scenes as well as static scenes.
We calculate the average of $k$ successive frames to generate a single blurry frame.
By averaging $k$ successive frames, realistic motion blurs from both moving objects and the camera shake can be rendered in the blurry frame and the $240/k$ fps blurry video can be generated (i.e. 16 fps video is generated by averaging every 15 frames). Notably, ground truth sharp frame is chosen to be the mid-frame used in averaging, since we aim to restore the latent frame captured in the middle of exposure time as shown in fig.~\ref{fig_kernel}. Thus the duty cycle is $\tau_i = 0.5$, in our whole dataset.
The videos are recorded with caution so that the motions should be no greater than 1 pixel between two neighboring frames to render more smooth and realistic blurry frame.



Our dataset mainly captured outdoor scenes to avoid flickering effect of fluorescent light which occurs when we capture indoor scenes with the high-speed camera. We captured numerous scenes in both dynamic and static environments, and each frame has HD (1280x720) size.
In Fig. \ref{fig_dataset_example}, some examples of our ground truth frames and rendered blurry frames are shown. We can see that the generated blurs are locally varying according to the depth changes and moving objects. Our dataset is publicly available on our website~\footnote{\url{http://cv.snu.ac.kr}}. We provide the generated blurry and corresponding sharp videos as well as the original videos we recorded.



\section{Experimental Results}
In this section, we empirically demonstrate the superiority of the proposed method over conventional methods.

\begin{figure*}[t]
	\begin{center}
		\includegraphics[width=\linewidth]{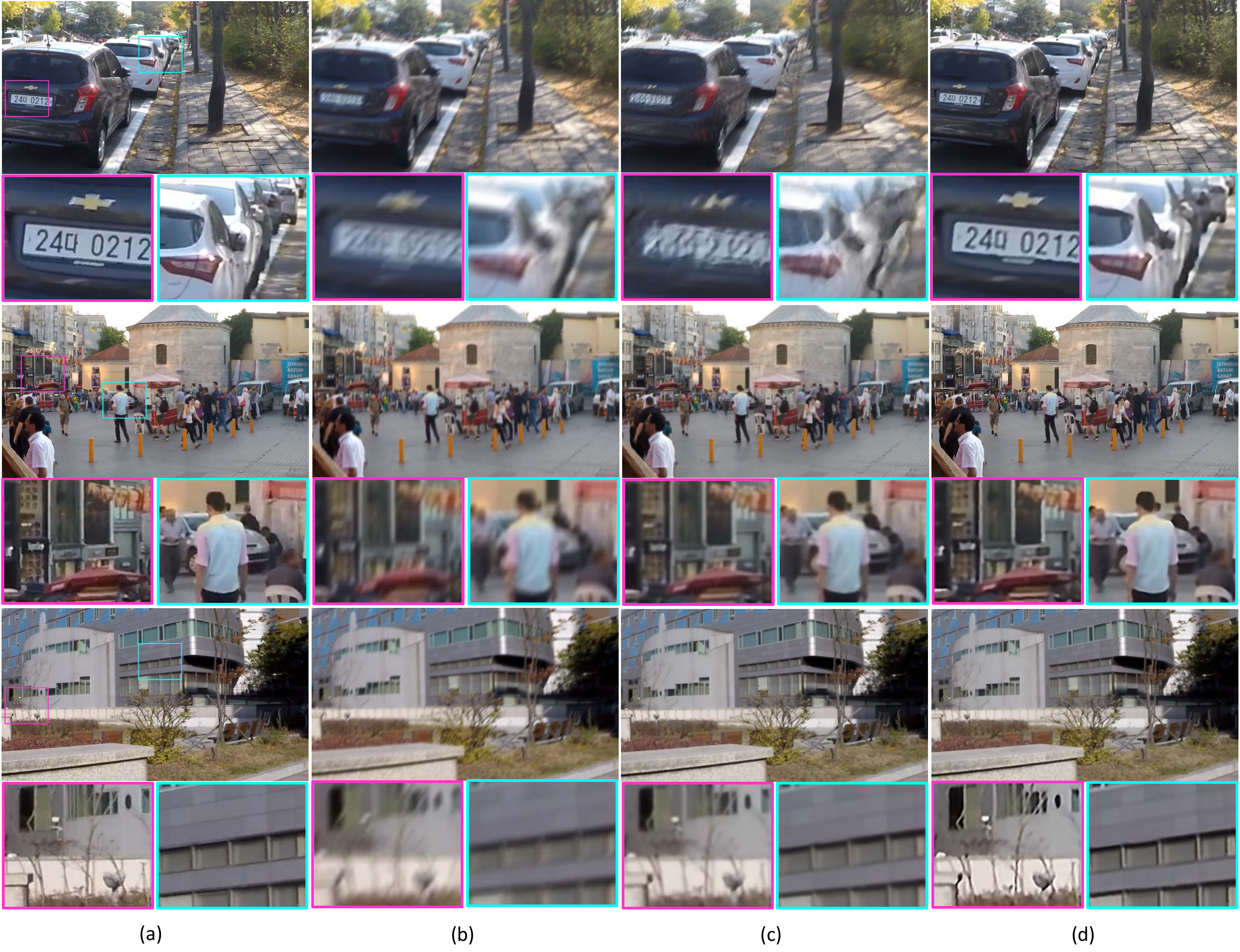}
	\end{center}
	\caption{Comparative deblurring results using our dataset. (a) Ground truth sharp frames. (b) Generated blurry frames. (c) Our results without considering defocus blurs. (d) Our final results.}
	\label{fig_experiment_success}
\end{figure*}

\begin{figure}[t]
	\begin{center}
		\includegraphics[width=\linewidth]{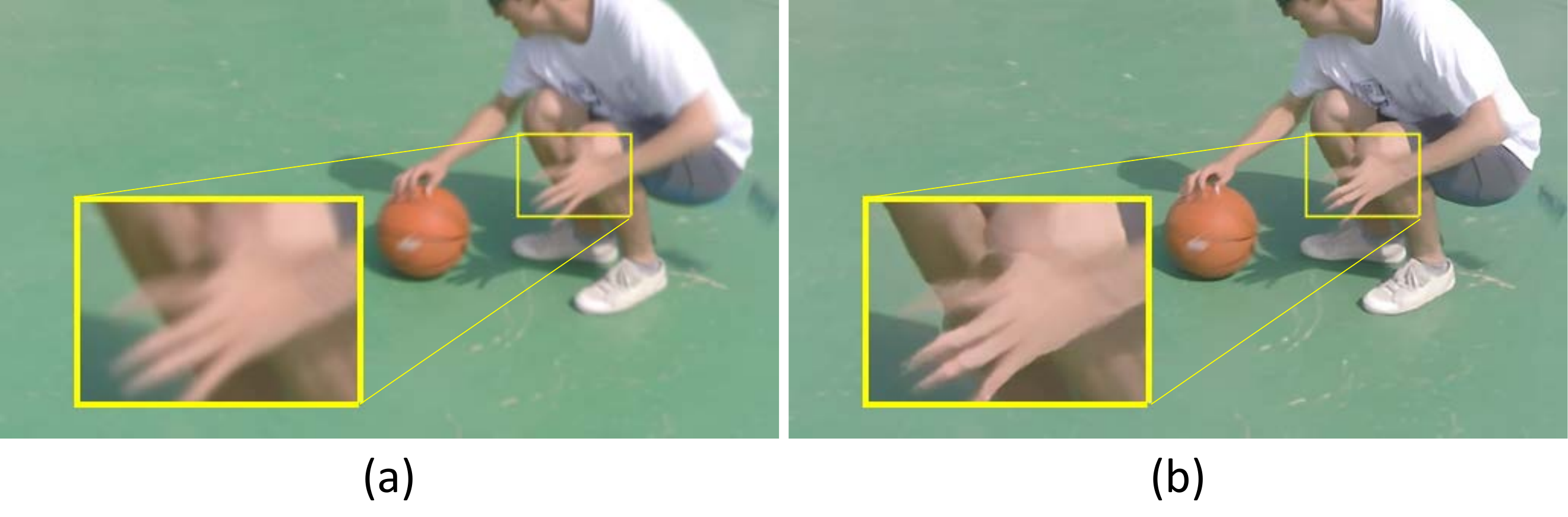}
	\end{center}
	\vspace{-3Ex}
	\caption{A failure case. (a) A blurry frame in the proposed dataset. (b) Our deblurring result.}
	\label{fig_experiment_failure}
\end{figure}

In Table~\ref{table_evaluation1} and Table~\ref{table_evaluation2}, our deblurring results are quantitatively evaluated with the proposed dataset.
For evaluation, we use fixed parameters and the values are $\lambda = 250$, $\mu = 2$, $\nu_{u} = \nu_{\sigma} = 0.08\lambda$, $v_I= (\frac{25}{255})^2$, and $N = 2$.
Since the source codes of other video deblurring methods that can handle non-uniform blur are not available, we evaluate our method in different settings. 
First, we calculate and compare both the PSNR and SSIM values of each original blurry sequence and the corresponding  deblurred one.
As our dataset contains only motion blurs in it, we restore the latent frames without considering the defocus blur (defocus blur kernel is set to be identity matrix).
Next, to demonstrate the good performance of the proposed method in removing defocus blurs,
we re-generate blurry dataset by adding Gaussian blur ($\sigma=1.5$) to the original sharp video before averaging. Using this dataset which contains both motion blur and defocus blur, we compare the our result against
each original blurry sequence and our deblurring result that does not consider defocus blur.
We verify that, our approach improves the deblurring results significantly in terms of PSNR and SSIM by removing blurs from defocus.
In Fig.~\ref{fig_experiment_success}, qualitative comparisons using our dataset are shown.
Ours restores the edges of buildings, letters, and moving persons, clearly.
However, we observe some failure cases in our results.
In Fig.~\ref{fig_experiment_failure}, we fail to estimate motions of fast moving hand, and thus 
fail in deblurring, since it is difficult to estimate accurate flows of small structure with distinct motions in the coarse-to-fine framework as reported in~\cite{Xu:2012}.

\begin{table}[h]
	\renewcommand{\arraystretch}{1.3}
	\caption{Deblurring performance evaluations in terms of PSNR.}
		\vspace{-3Ex}
	\label{table_evaluation1}
	\begin{center}
		\begin{tabular}{|l||c c|c c c|}
			\hline
			& \multicolumn{2}{c|}{Motion blur only} & \multicolumn{3}{|c|}{Motion blur + Gaussian blur ($\sigma$ = 1.5)} \\
			\bfseries Seq.  & \bfseries Blurry & \bfseries ours & \bfseries Blurry & \bfseries ours  & \bfseries ours\\
			&  & (w/o defocus) &  &(w/o defocus) &(full)\\
			\hline\hline
			
			\#1 & 26.8 & 27.79 & 25.88 & 27.53 & 27.67\\
			\#2 & 26.5 & 27.68 & 24.29 & 25.18 & 26.18\\
			\#3 & 33.28 & 34.78 & 30.55 & 31.27 & 32.65\\
			\#4 & 37.07 & 36.94 & 36.52 & 36.50 & 36.36\\
			\#5 & 24.34 & 23.62 & 23.78 & 23.05 & 24.24\\
			\#6 & 26.83 & 29.07 & 24.04 & 25.18 & 26.46\\
			\#7 & 29.03 & 30.52 & 25.95 & 27.31 & 28.55\\
			\#8 & 24.80 & 29.81 & 23.57 & 26.05 & 27.01\\
			\#9 & 28.55 & 31.41 & 27.19 & 29.05 & 27.74\\
			\#10 & 26.13 & 30.55 & 24.83 & 27.61 & 28.25\\
			\#11 & 29.24 & 33.61 & 27.73 & 30.47 & 30.86\\

			\hline\hline
			\bfseries Avg. & 28.42 & 30.52 & 26.76 & 28.11 & 28.73\\
			\hline
		\end{tabular}
	\end{center}
%
%
%
%
	\renewcommand{\arraystretch}{1.3}
	\caption{Deblurring performance evaluations in terms of SSIM.}
	\vspace{-3Ex}
	\label{table_evaluation2}
	\begin{center}
		\begin{tabular}{|l||c c|c c c|}
			\hline
			& \multicolumn{2}{c|}{Motion blur only} & \multicolumn{3}{|c|}{Motion blur + Gaussian blur ($\sigma$ = 1.5)} \\
			\bfseries Seq.  & \bfseries Blurry & \bfseries ours & \bfseries Blurry & \bfseries ours  & \bfseries ours\\
			&  & (w/o defocus) &  &(w/o defocus) &(full)\\
			\hline\hline

			\#1 & 0.8212 & 0.8611 & 0.7898 & 0.8374 & 0.8476\\
			\#2 & 0.8571 & 0.8847 & 0.7526 & 0.7809 & 0.8197\\
			\#3 & 0.9327 & 0.9473 & 0.8750 & 0.8849 & 0.9087\\
			\#4 & 0.9701 & 0.9695 & 0.9652 & 0.9665 & 0.9657\\
			\#5 & 0.7154 & 0.7181 & 0.6853 & 0.6598 & 0.7293\\
			\#6 & 0.8362 & 0.9178 & 0.7362 & 0.7880 & 0.8334\\
			\#7 & 0.8751 & 0.9244 & 0.7928 & 0.8360 & 0.8694\\
			\#8 & 0.8068 & 0.9269 & 0.7529 & 0.8320 & 0.8582\\
			\#9 & 0.8322 & 0.9100 & 0.7908 & 0.8427 & 0.8854\\
			\#10 & 0.8083 & 0.9198 & 0.7620 & 0.8432 & 0.8645\\
			\#11 & 0.9176 & 0.9608 & 0.8945 & 0.9283 & 0.9367\\

			\hline\hline
			\bfseries Avg. & 0.8521 & 0.9037 & 0.7997 & 0.8363 & 0.9367\\
			\hline
		\end{tabular}
	\end{center}
	\vspace{-3Ex}
\end{table}

\begin{figure*}[t]
	\begin{center}
		\includegraphics[width=\linewidth]{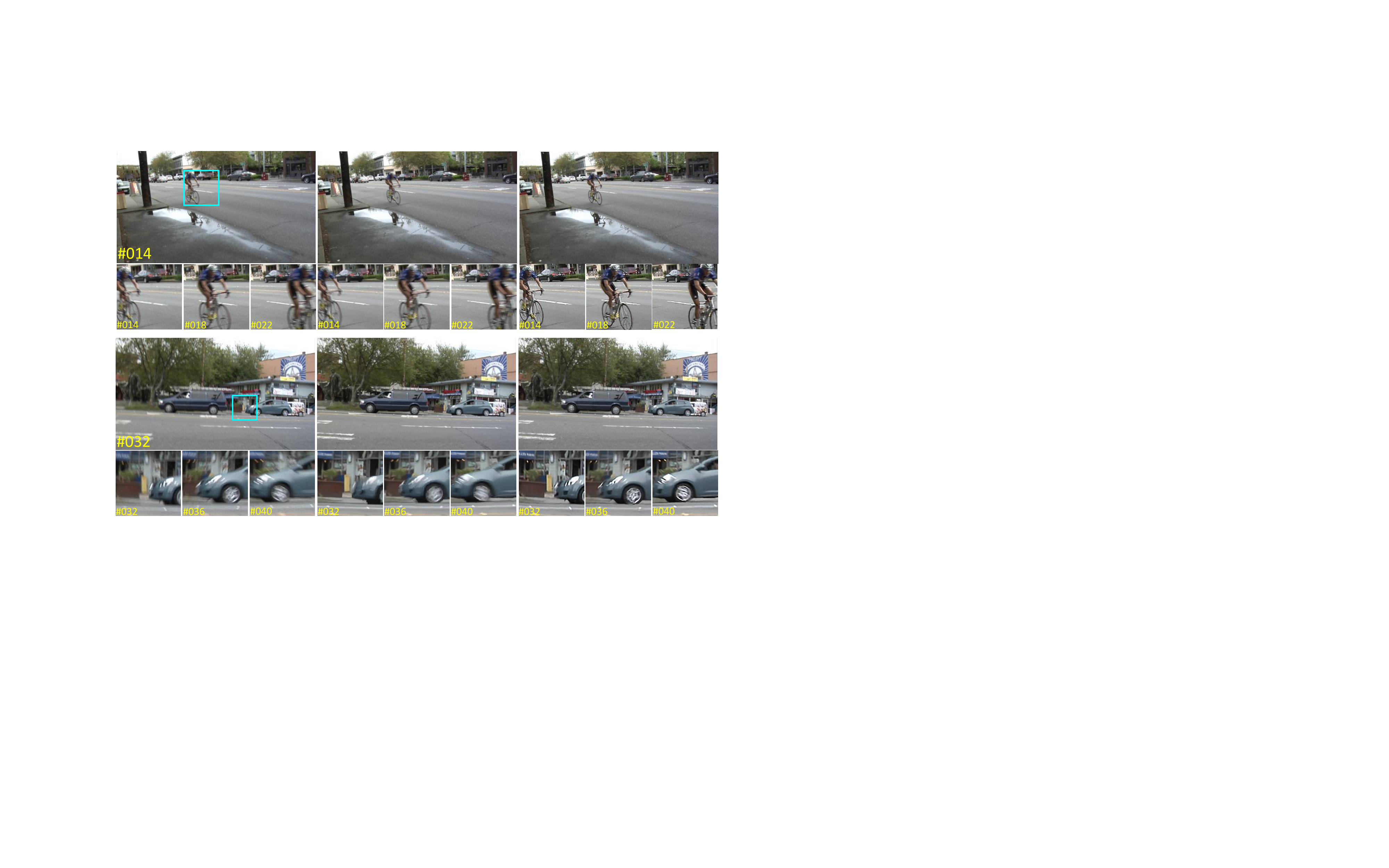}
	\end{center}
	\caption{\textbf{Left to right:} Blurry frames of dynamic scenes, deblurring results of \cite{cho_siggraph2012}, and our results.}
	\label{fig_comp_exemplar1}
%
%
	
	\begin{center}
		\includegraphics[width=\linewidth]{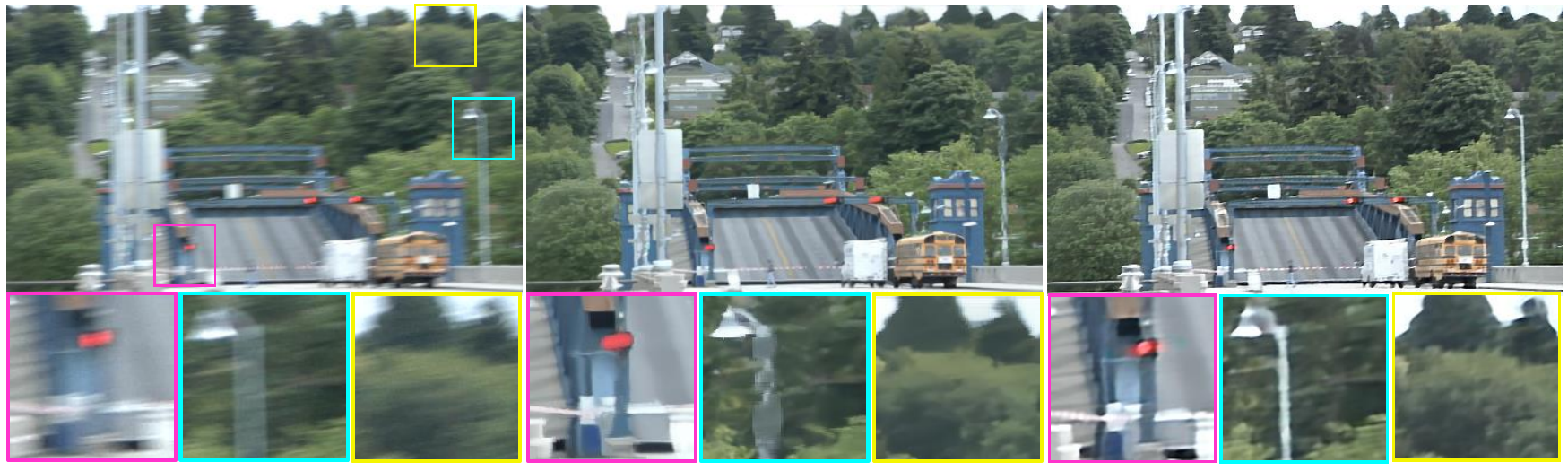}
	\end{center}
	\caption{\textbf{Left to right:} Blurry frame, deblurring result of \cite{cho_siggraph2012}, and ours.}
	\label{fig_comp_exemplar2}

\end{figure*}

\begin{figure*}[t]
	\begin{center}
		\includegraphics[width=\linewidth]{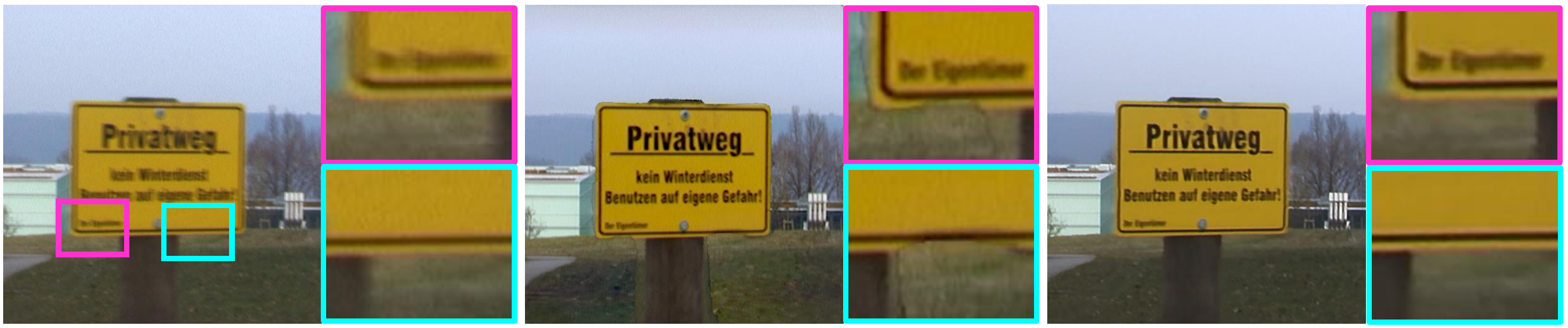}
	\end{center}
	\caption{Comparison with segmentation-based approach. \textbf{Left to right:} Blurry frame, result of \cite{Wulff:ECCV:2014}, and ours.}
	\label{fig_com_segmentation}	
\end{figure*}

Next, we compare our deblurring results with 
those of the state-of-the art exemplar based method \cite{cho_siggraph2012} 
with the videos used in \cite{cho_siggraph2012}.
As shown in Fig. \ref{fig_comp_exemplar1}, the captured scenes are dynamic and contain multiple moving objects.
The method \cite{cho_siggraph2012} fails in restoring the moving objects,
because the object motions are large and distinct from the backgrounds.
By contrast, our results show better performances in deblurring moving objects and backgrounds.
Notably, the exemplar-based approach also fails in handling large blurs, as shown in Fig. \ref{fig_comp_exemplar2},
as the initially estimated homographies in the largely blurred images are inaccurate.
Moreover, this approach renders excessively smooth results for mid-frequency textures such as trees,
as the method is based on interpolation without spatial prior for latent frames.

We also compare our method with the state-of-the-art segmentation-based approach \cite{Wulff:ECCV:2014}.
The test video is shown in Fig.~\ref{fig_com_segmentation}, which is a bilayer scene used in \cite{Wulff:ECCV:2014}.
Although the bi-layer scene is a good example to verify the performance of the layered model,
inaccurate segmentation near the boundaries causes serious artifacts in the restored frame.
By contrast, since our method does not need segmentation
and it restores the boundaries much better than the layered model.

\begin{figure}[h]
	\begin{center}
		\includegraphics[width=\linewidth]{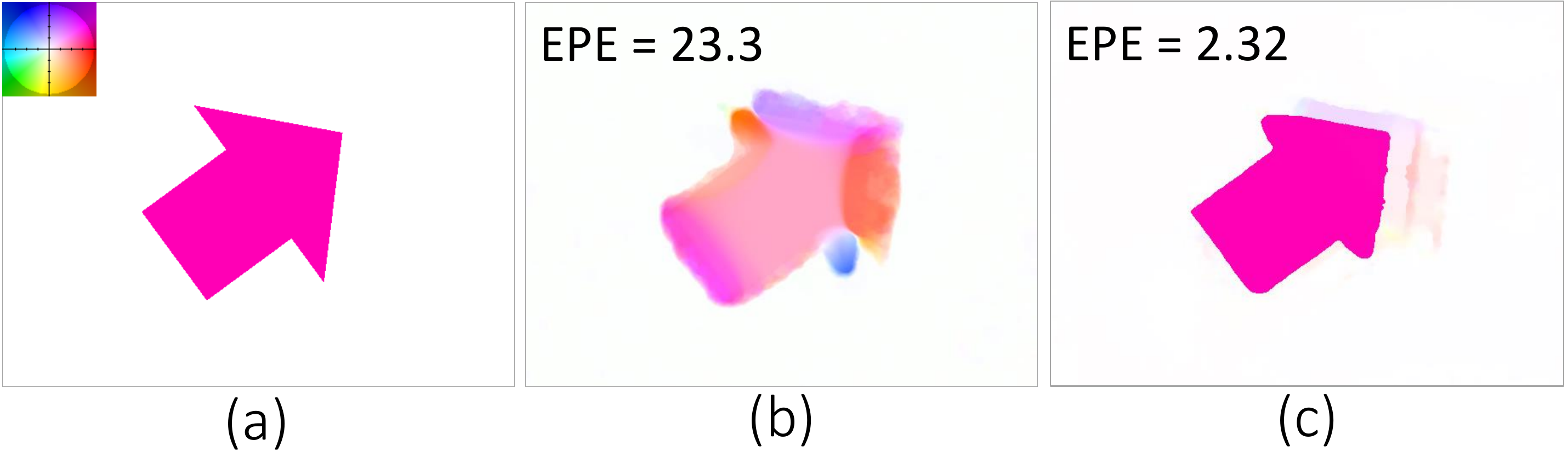}
	\end{center}
	\caption{EPE denotes average end point error. (a) Color coded ground truth optical flow between blurry images. (b) Optical flow estimation result of \cite{Portz:2012}. (c) Our result.}
	\label{fig_comp_optical_flow}
%
	
	\begin{center}
		\includegraphics[width=\linewidth]{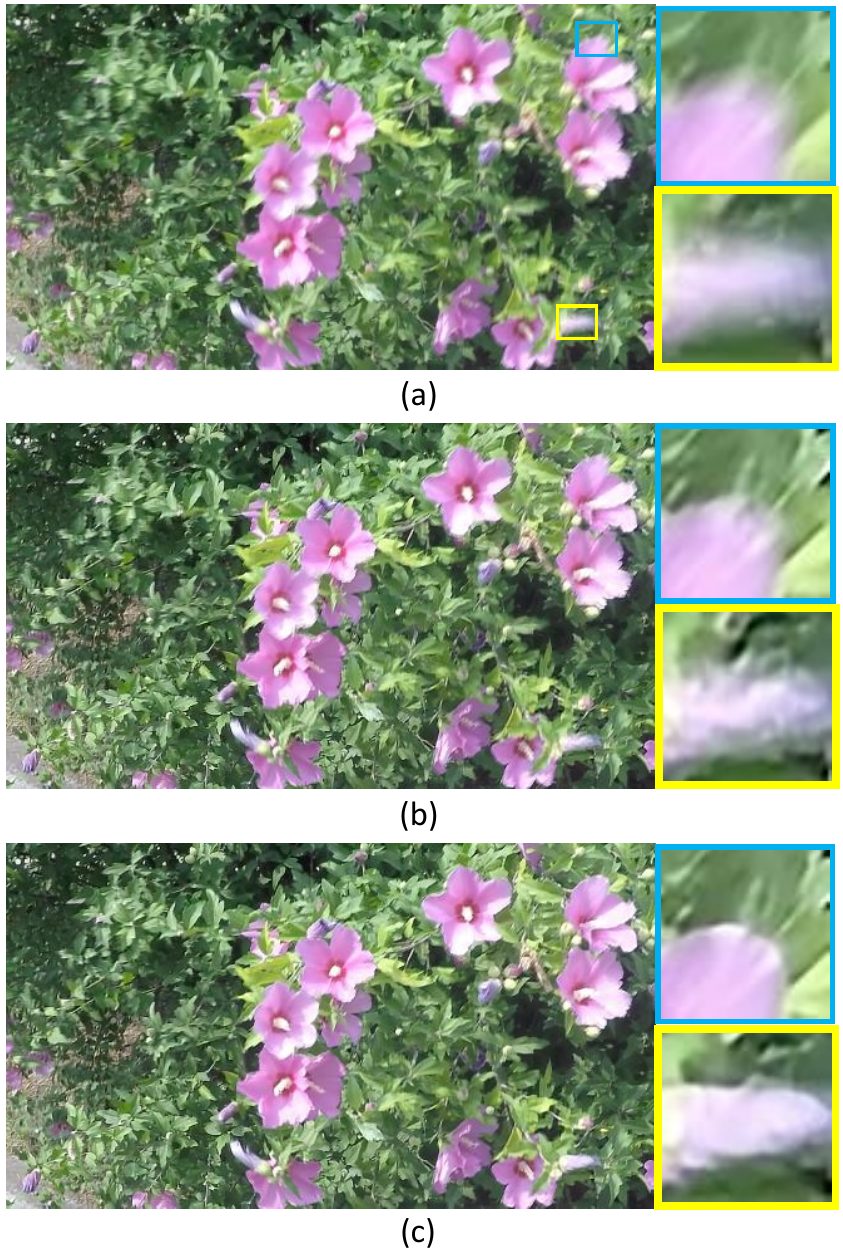}
	\end{center}
	\caption{(a) A blurry frame of a video. (b) Our deblurring result without using $\textbf{E}_{temporal}$. (c) Our deblurring result with $\textbf{E}_{temporal}$.}
	\label{fig_temporal}	
	
	\vspace{-3Ex}
\end{figure}

In Fig.~\ref{fig_comp_optical_flow}, we quantitatively compare the optical flow accuracies with \cite{Portz:2012}
on synthetic blurry images. As publicly available code of~\cite{Portz:2012} cannot handle Gaussian blur,
we synthesize blurry frames which have motion blurs only.
Although \cite{Portz:2012} was proposed to handle blurry images in optical flow estimation,
its assumption does not hold in motion boundaries, which is very important for deblurring.
Therefore, their optical flow is inaccurate in the motion boundaries of moving objects.
However, our model can cope with abrupt motion changes, and thus performs better than the previous models.

Moreover, we show the deblurring results
with and without using the temporal coherence term in (\ref{equ_temporal}),
and verify that our temporal coherence model clearly restores edges and significantly reduces ringing artifacts near the edges in Fig.~\ref{fig_temporal}.

\begin{figure}[t]
	\begin{center}
		\includegraphics[width=\linewidth]{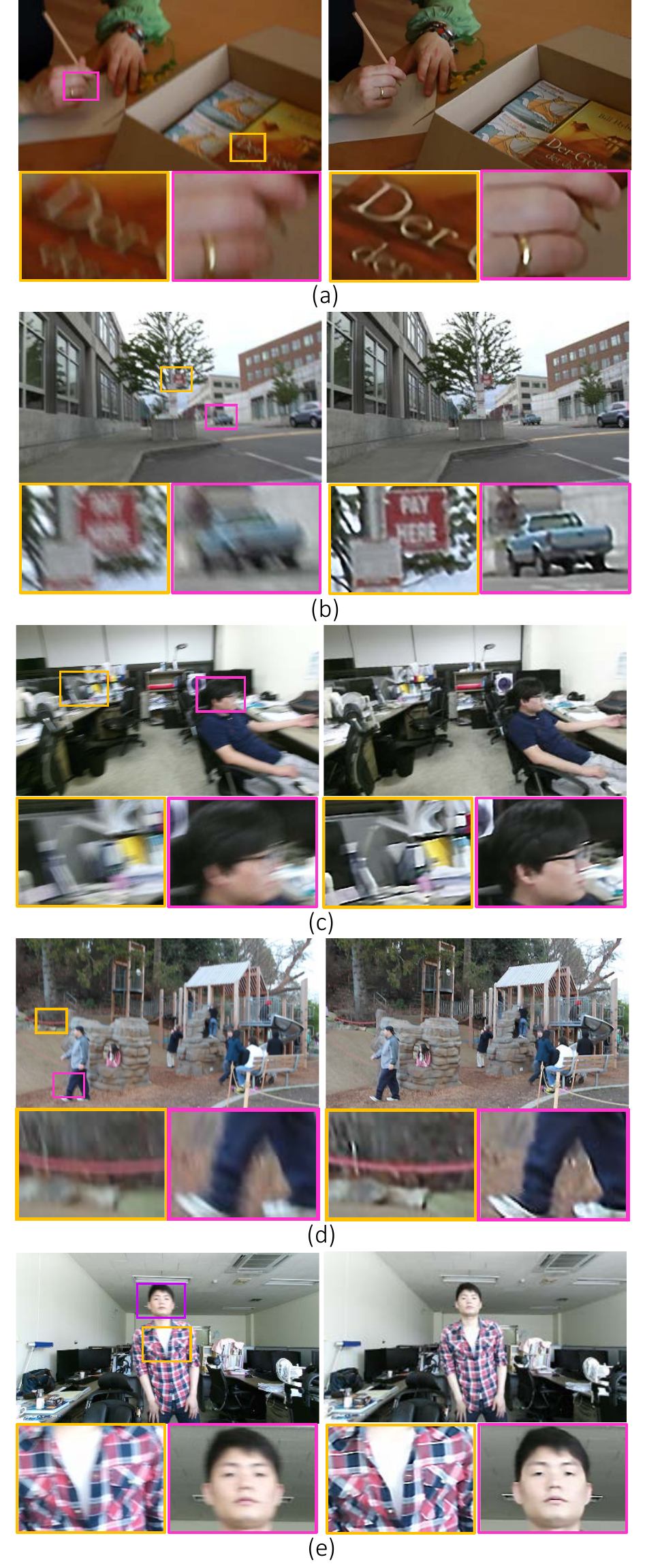}
	\end{center}
	\caption{\textbf{Left to right:} Numerous real blurry frames and our deblurring results.}
	\label{fig_comp_self}	
\end{figure}

Finally, other deblurring results from numerous real videos are shown in Fig.~\ref{fig_comp_self}.
Notably, our model successfully restores the face which has highly non-uniform blurs 
because the person moves rotationally (Fig.~\ref{fig_comp_self}(e)).

The video demo is provided in the supplementary material.
For additional results, please see the supplementary material.




\section{Conclusion}
In this study, we introduced a novel method that removes general blurs in dynamic scenes
which conventional methods fail to.
We inferred bidirectional optical flows to approximate motion blur kernels,
and estimated the scales of Gaussian blurs to approximate defocus blur kernels.
Therefore we handled general blurs, by estimating a pixel-wise different blur kernel.
In addition, we proposed a new single energy model that estimates optical flows, defocus blur maps and latent frames, jointly.
We also provided a framework and efficient solvers to minimize the proposed energy function
and it has been shown that our method yields superior deblurring results to several
state-of-the-art deblurring methods through intensive experiments with real challenging blurred videos. 
Moreover, we provided the publicly available benchmark dataset to evaluate the non-uniform deblurring methods and we quantitatively evaluated the performance of the proposed method using the proposed dataset.
Nevertheless, our model has its limitations in handling large displacement fields.
Therefore, improving the proposed algorithm to handle large displacements is required.
Moreover, since our current work is implemented on Matlab, it is time consuming and needs large resources. 
Thus, for practical applications, reducing the running time by code optimization and parallel implementation as well as efficient memory management will be considered in our future work.

%

%
%

\ifCLASSOPTIONcaptionsoff
\newpage
\fi


\bibliographystyle{IEEEtran}
\bibliography{VD}

\begin{thebibliography}{10}
\providecommand{\url}[1]{#1}
\csname url@samestyle\endcsname
\providecommand{\newblock}{\relax}
\providecommand{\bibinfo}[2]{#2}
\providecommand{\BIBentrySTDinterwordspacing}{\spaceskip=0pt\relax}
\providecommand{\BIBentryALTinterwordstretchfactor}{4}
\providecommand{\BIBentryALTinterwordspacing}{\spaceskip=\fontdimen2\font plus
\BIBentryALTinterwordstretchfactor\fontdimen3\font minus
  \fontdimen4\font\relax}
\providecommand{\BIBforeignlanguage}[2]{{%
\expandafter\ifx\csname l@#1\endcsname\relax
\typeout{** WARNING: IEEEtran.bst: No hyphenation pattern has been}%
\typeout{** loaded for the language `#1'. Using the pattern for}%
\typeout{** the default language instead.}%
\else
\language=\csname l@#1\endcsname
\fi
#2}}
\providecommand{\BIBdecl}{\relax}
\BIBdecl

\bibitem{Cho:2009}
S.~Cho and S.~Lee, ``Fast motion deblurring,'' in \emph{SIGGRAPH}, 2009.

\bibitem{Fergus:2006}
R.~Fergus, B.~Singh, A.~Hertzmann, S.~T. Roweis, and W.~Freeman, ``Removing
  camera shake from a single photograph,'' in \emph{SIGGRAPH}, 2006.

\bibitem{Gupta:2010}
A.~Gupta, N.~Joshi, L.~Zitnick, M.~Cohen, and B.~Curless, ``Single image
  deblurring using motion density functions,'' in \emph{ECCV}, 2010.

\bibitem{Hirsch:2011}
M.~Hirsch, C.~J. Schuler, S.~Harmeling, and B.~Scholkopf, ``Fast removal of
  non-uniform camera shake,'' in \emph{Computer Vision (ICCV), 2011 IEEE
  International Conference on}.\hskip 1em plus 0.5em minus 0.4em\relax IEEE,
  2011, pp. 463--470.

\bibitem{Shan:2008}
Q.~Shan, J.~Jia, and A.~Agarwala, ``High-quality motion deblurring from a
  single image,'' in \emph{SIGGRAPH}, 2008.

\bibitem{Whyte:2012}
O.~Whyte, J.~Sivic, A.~Zisserman, and J.~Ponce, ``Non-uniform deblurring for
  shaken images,'' \emph{International Journal of Computer Vision}, vol.~98,
  no.~2, pp. 168--186, 2012.

\bibitem{cai2009blind}
J.-F. Cai, H.~Ji, C.~Liu, and Z.~Shen, ``Blind motion deblurring using multiple
  images,'' \emph{Journal of computational physics}, vol. 228, no.~14, pp.
  5057--5071, 2009.

\bibitem{li2010generating}
Y.~Li, S.~B. Kang, N.~Joshi, S.~M. Seitz, and D.~P. Huttenlocher, ``Generating
  sharp panoramas from motion-blurred videos,'' in \emph{Proc. IEEE
  International Conference on Computer Vision and Pattern Recognition}, 2010.

\bibitem{tai2011richardson}
Y.-W. Tai, P.~Tan, and M.~S. Brown, ``Richardson-lucy deblurring for scenes
  under a projective motion path,'' \emph{Pattern Analysis and Machine
  Intelligence, IEEE Transactions on}, vol.~33, no.~8, pp. 1603--1618, 2011.

\bibitem{cho2012registration}
S.~Cho, H.~Cho, Y.-W. Tai, and S.~Lee, ``Registration based non-uniform motion
  deblurring,'' in \emph{Computer Graphics Forum}, vol.~31, no.~7.\hskip 1em
  plus 0.5em minus 0.4em\relax Wiley Online Library, 2012, pp. 2183--2192.

\bibitem{paramanand2013non}
C.~Paramanand and A.~N. Rajagopalan, ``Non-uniform motion deblurring for
  bilayer scenes,'' in \emph{Proc. IEEE International Conference on Computer
  Vision and Pattern Recognition}, 2013.

\bibitem{lee2013dense}
H.~S. Lee and K.~M. Lee, ``Dense 3d reconstruction from severely blurred images
  using a single moving camera,'' in \emph{Proc. IEEE International Conference
  on Computer Vision and Pattern Recognition}, 2013.

\bibitem{cho2007removing}
S.~Cho, Y.~Matsushita, and S.~Lee, ``Removing non-uniform motion blur from
  images,'' in \emph{Computer Vision, 2007. ICCV 2007. IEEE 11th International
  Conference on}.\hskip 1em plus 0.5em minus 0.4em\relax IEEE, 2007, pp. 1--8.

\bibitem{bar2007variational}
L.~Bar, B.~Berkels, M.~Rumpf, and G.~Sapiro, ``A variational framework for
  simultaneous motion estimation and restoration of motion-blurred video,'' in
  \emph{Proc. IEEE International Conference on Computer Vision and Pattern
  Recognition}, 2007.

\bibitem{Wulff:ECCV:2014}
J.~Wulff and M.~J. Black, ``Modeling blurred video with layers,'' in
  \emph{ECCV}, 2014.

\bibitem{thkim_cvpr2014}
T.~H. Kim and K.~M. Lee, ``Segmentation-free dynamic scene deblurring,'' in
  \emph{Proc. IEEE International Conference on Computer Vision and Pattern
  Recognition}, 2014.

\bibitem{matsushita2006full}
Y.~Matsushita, E.~Ofek, W.~Ge, X.~Tang, and H.-Y. Shum, ``Full-frame video
  stabilization with motion inpainting,'' \emph{Pattern Analysis and Machine
  Intelligence, IEEE Transactions on}, vol.~28, no.~7, pp. 1150--1163, 2006.

\bibitem{cho_siggraph2012}
S.~Cho, J.~Wang, and S.~Lee, ``Video deblurring for hand-held cameras using
  patch-based synthesis,'' \emph{ACM Transactions on Graphics}, vol.~31, no.~4,
  pp. 64:1--64:9, 2012.

\bibitem{bae2007defocus}
S.~Bae and F.~Durand, ``Defocus magnification,'' in \emph{Computer Graphics
  Forum}, vol.~26, no.~3.\hskip 1em plus 0.5em minus 0.4em\relax Wiley Online
  Library, 2007, pp. 571--579.

\bibitem{kee2011modeling}
E.~Kee, S.~Paris, S.~Chen, and J.~Wang, ``Modeling and removing
  spatially-varying optical blur,'' in \emph{Computational Photography (ICCP),
  2011 IEEE International Conference on}.\hskip 1em plus 0.5em minus
  0.4em\relax IEEE, 2011, pp. 1--8.

\bibitem{zhu2013estimating}
X.~Zhu, S.~Cohen, S.~Schiller, and P.~Milanfar, ``Estimating spatially varying
  defocus blur from a single image,'' \emph{Image Processing, IEEE Transactions
  on}, vol.~22, no.~12, pp. 4879--4891, 2013.

\bibitem{zhuo2011defocus}
S.~Zhuo and T.~Sim, ``Defocus map estimation from a single image,''
  \emph{Pattern Recognition}, vol.~44, no.~9, pp. 1852--1858, 2011.

\bibitem{thkim_cvpr2015}
T.~Hyun~Kim and K.~Mu~Lee, ``Generalized video deblurring for dynamic scenes,''
  in \emph{Proceedings of the IEEE Conference on Computer Vision and Pattern
  Recognition}, 2015, pp. 5426--5434.

\bibitem{Levin:2009}
A.~Levin, Y.~Weiss, F.~Durand, and W.~T. Freeman, ``Understanding and
  evaluating blind deconvolution algorithms,'' in \emph{Proc. IEEE
  International Conference on Computer Vision and Pattern Recognition}, 2009.

\bibitem{kohler2012recording}
R.~K{\"o}hler, M.~Hirsch, B.~Mohler, B.~Sch{\"o}lkopf, and S.~Harmeling,
  ``Recording and playback of camera shake: Benchmarking blind deconvolution
  with a real-world database,'' in \emph{Computer Vision--ECCV 2012}.\hskip 1em
  plus 0.5em minus 0.4em\relax Springer, 2012, pp. 27--40.

\bibitem{xu2014deep}
L.~Xu, J.~S. Ren, C.~Liu, and J.~Jia, ``Deep convolutional neural network for
  image deconvolution,'' in \emph{Advances in Neural Information Processing
  Systems}, 2014, pp. 1790--1798.

\bibitem{Schuler_PAMI15}
C.~J. Schuler, M.~Hirsch, S.~Harmeling, and B.~Sch{\"o}lkopf, ``Learning to
  deblur,'' \emph{IEEE Transactions on Pattern Analysis and Machine
  Intelligence (PAMI)}, 2015.

\bibitem{sun2015learning}
J.~Sun, W.~Cao, Z.~Xu, and J.~Ponce, ``Learning a convolutional neural network
  for non-uniform motion blur removal,'' \emph{arXiv preprint
  arXiv:1503.00593}, 2015.

\bibitem{Dai:2008}
S.~Dai and Y.~Wu, ``Motion from blur,'' in \emph{Proc. IEEE International
  Conference on Computer Vision and Pattern Recognition}, 2008.

\bibitem{Portz:2012}
T.~Portz, L.~Zhang, and H.~Jiang, ``Optical flow in the presence of
  spatially-varying motion blur,'' in \emph{Proc. IEEE International Conference
  on Computer Vision and Pattern Recognition}, 2012.

\bibitem{shi2015just}
J.~Shi, L.~Xu, and J.~Jia, ``Just noticeable defocus blur detection and
  estimation,'' in \emph{Proceedings of the IEEE Conference on Computer Vision
  and Pattern Recognition}, 2015, pp. 657--665.

\bibitem{Krishnan:2009}
D.~Krishnan and R.~Fergus, ``Fast image deconvolution using hyper-laplacian
  priors,'' in \emph{NIPS}, 2009.

\bibitem{Krishnan:2011}
D.~Krishnan, T.~Tay, and R.~Fergus, ``Blind deconvolution using a normalized
  sparsity measure,'' in \emph{Proc. IEEE International Conference on Computer
  Vision and Pattern Recognition}, 2009.

\bibitem{Levin:PAMI2007}
A.~Levin and Y.~Weiss, ``User assisted separation of reflections from a single
  image using a sparsity prior,'' \emph{IEEE Trans. Pattern Analysis Machine
  Intelligence}, vol.~29, no.~9, pp. 1647--1654, 2007.

\bibitem{hu_cvpr2014_depthdeblur}
Z.~Hu, L.~Xu, and M.-H. Yang, ``Joint depth estimation and camera shake removal
  from single blurry image,'' in \emph{Proc. IEEE International Conference on
  Computer Vision and Pattern Recognition}, 2014.

\bibitem{thkim:2013}
T.~H. Kim, B.~Ahn, and K.~M. Lee, ``Dynamic scene deblurring,'' in
  \emph{Computer Vision (ICCV), 2013 IEEE International Conference on}.\hskip
  1em plus 0.5em minus 0.4em\relax IEEE, 2013, pp. 3160--3167.

\bibitem{thkim_iccv2013optical}
T.~H. Kim, H.~S. Lee, and K.~M. Lee, ``Optical flow via locally adaptive fusion
  of complementary data costs,'' in \emph{Computer Vision (ICCV), 2013 IEEE
  International Conference on}.\hskip 1em plus 0.5em minus 0.4em\relax IEEE,
  2013, pp. 3344--3351.

\bibitem{Chambolle:2011}
\BIBentryALTinterwordspacing
A.~Chambolle and T.~Pock, ``A first-order primal-dual algorithm for convex
  problems with applications to imaging,'' \emph{Journal of Mathematical
  Imaging and Vision}, vol.~40, no.~1, pp. 120--145, May 2011. [Online].
  Available: \url{http://dx.doi.org/10.1007/s10851-010-0251-1}
\BIBentrySTDinterwordspacing

\bibitem{wedel2009improved}
A.~Wedel, T.~Pock, C.~Zach, H.~Bischof, and D.~Cremers, ``An improved algorithm
  for tv-l 1 optical flow,'' in \emph{Statistical and Geometrical Approaches to
  Visual Motion Analysis}.\hskip 1em plus 0.5em minus 0.4em\relax Springer,
  2009, pp. 23--45.

\bibitem{rhemann2011fast}
C.~Rhemann, A.~Hosni, M.~Bleyer, C.~Rother, and M.~Gelautz, ``Fast cost-volume
  filtering for visual correspondence and beyond,'' in \emph{Computer Vision
  and Pattern Recognition (CVPR), 2011 IEEE Conference on}.\hskip 1em plus
  0.5em minus 0.4em\relax IEEE, 2011, pp. 3017--3024.

\bibitem{agrawal2009optimal}
A.~Agrawal and R.~Raskar, ``Optimal single image capture for motion
  deblurring,'' in \emph{Computer Vision and Pattern Recognition, 2009. CVPR
  2009. IEEE Conference on}.\hskip 1em plus 0.5em minus 0.4em\relax IEEE, 2009,
  pp. 2560--2567.

\bibitem{Xu:2012}
L.~Xu, J.~Jia, and Y.~Matsushita, ``Motion detail preserving optical flow
  estimation,'' \emph{IEEE Trans. Pattern Analysis Machine Intelligence},
  vol.~34, no.~9, pp. 1744--1757, 2012.

\end{thebibliography}

%

\clearpage
\begin{IEEEbiography}[{\includegraphics[width=1in,height=1.25in,clip,keepaspectratio]{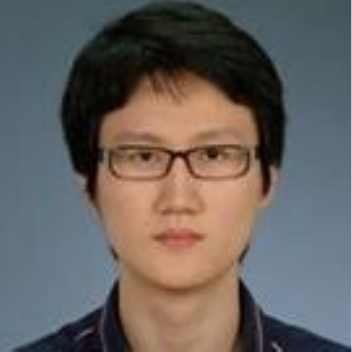}}]{Tae Hyun Kim}
received the BS degree and the MS degree in the department of electrical
engineering from KAIST, Daejeon, Korea, in 2008
and 2010, respectively. He is currently working
toward the PhD degree in Electrical and Computer Engineering at Seoul National University.
His research interests include motion estimation, and deblurring. He is a
student member of the IEEE.
\end{IEEEbiography}

\begin{IEEEbiography}[{\includegraphics[width=1in,height=1.25in,clip,keepaspectratio]{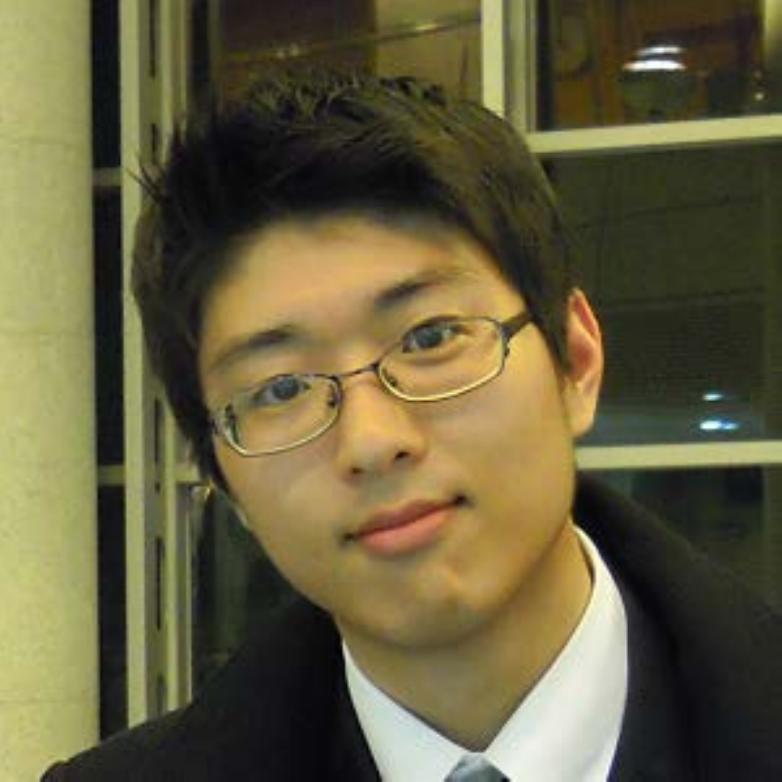}}]{Seungjun Nah}

received the BS degree in Electrical and Computer Engineering from Seoul National University (SNU), Seoul, Korea in 2014. He is currently working towards PhD degree in Electrical and Computer Engineering at Seoul National University. He is interested in computer vision problems including deblurring and visual saliency. He is a student member of the IEEE.

\end{IEEEbiography}


\begin{IEEEbiography}[{\includegraphics[width=1in,height=1.25in,clip,keepaspectratio]{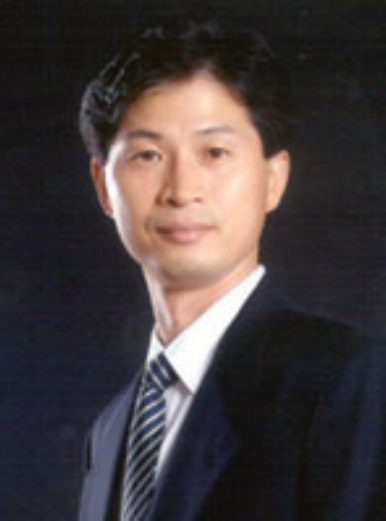}}]{Kyoung Mu Lee}

received the BS and MS
degrees in control and instrumentation engineering
from Seoul National University (SNU), Seoul,
Korea in 1984 and 1986, respectively, and the
PhD degree in electrical engineering from the
University of Southern California (USC), Los
Angeles, California in 1993. He received the
Korean Government Overseas Scholarship during
the PhD courses. From 1993 to 1994, he was
a research associate in the Signal and Image
Processing Institute (SIPI) at USC. He was with
the Samsung Electronics Co. Ltd. in Korea as a senior researcher from
1994 to 1995. In August 1995, he joined the Department of Electronics
and Electrical Engineering of the Hong-Ik University, and was an assistant
and associate professor. Since September 2003, he has been with
the Department of Electrical and Computer Engineering at
Seoul National University as a professor, and leads the Computer Vision
Laboratory. His primary research is focused on statistical methods in
computer vision that can be applied to various applications including
object recognition, segmentation, tracking and 3D reconstruction. He
has received several awards, in particular, the Most Influential Paper
over the Decade Award by the IAPR Machine Vision Application in 2009,
the ACCV Honorable Mention Award in 2007, the Okawa Foundation
Research Grant Award in 2006, and the Outstanding Research Award
by the College of Engineering of SNU in 2010. He served as an Associate Editor in Chief, Editorial
Board member of the EURASIP Journal of Applied Signal Processing,
and is an associate editor of the Machine Vision Application Journal, the
IPSJ Transactions on Computer Vision and Applications, and the Journal
of Information Hiding and Multimedia Signal Processing. He has
(co)authored more than 100 publications in refereed journals and conferences
including PAMI, IJCV, CVPR, ICCV, and ECCV.

\end{IEEEbiography}




\end{document}